\documentclass[lettersize,journal]{IEEEtran}
\usepackage{graphicx} 
\usepackage{subcaption} 
\usepackage{multirow}
\usepackage{float}
\usepackage{booktabs}
\usepackage{amsmath}
\usepackage{stfloats} 
\usepackage{comment}
\usepackage{cite}
\usepackage{amssymb}  
\usepackage{textcomp}
\def\BibTeX{{\rm B\kern-.05em{\sc i\kern-.025em b}\kern-.08em
		T\kern-.1667em\lower.7ex\hbox{E}\kern-.125emX}}

\usepackage[pagebackref=true,breaklinks=true,colorlinks=true,bookmarksopen=true,bookmarksnumbered=true,citecolor=blue,urlcolor=blue]{hyperref}

\title{\LARGE \bf
{MA-SLAM: Active SLAM in Large-Scale Unknown Environment using Map Aware Deep Reinforcement Learning}
}

\author{Yizhen Yin, Yuhua Qi, Dapeng Feng, Hongbo Chen, \\ Hongjun Ma, \IEEEmembership{Member, IEEE}, Jin Wu, \IEEEmembership{Member, IEEE}, and Yi Jiang \IEEEmembership{Member, IEEE}

\thanks{Corresponding author: Yuhua Qi}
\thanks{Yizhen Yin, Yuhua Qi, Dapeng Feng and Hongbo Chen
    	are with Sun Yat-sen University, Guangzhou, China (e-mail: \{yinyzh5,fengdp5\}@mail2.sysu.edu.cn, \{qiyh8, chenhongbo\}@mail.sysu.edu.cn).}
\thanks{Hongjun Ma is with South China University of Technology, Guangzhou, China (e-mail: mahongjun@scut.edu.cn).}
\thanks{Jin Wu is with Hong Kong University of Science and Technology, Hong Kong, China (e-mail: jin\_wu\_uestc@hotmail.com).}
\thanks{Yi Jiang is with City University of Hong Kong, Hong Kong, China (e-mail: yjian22@cityu.edu.hk).}
}

\begin{document}
\maketitle

\begin{abstract}
Active Simultaneous Localization and Mapping (Active SLAM) involves the strategic planning and precise control of a robotic system's movement in order to construct a highly accurate and comprehensive representation of its surrounding environment, which has garnered significant attention within the research community. While the current methods demonstrate efficacy in small and controlled settings, they face challenges when applied to large-scale and diverse environments, marked by extended periods of exploration and suboptimal paths of discovery. In this paper, we propose MA-SLAM, a Map-Aware Active SLAM system based on Deep Reinforcement Learning (DRL), designed to address the challenge of efficient exploration in large-scale environments. In pursuit of this objective, we put forward a novel structured map representation. By discretizing the spatial data and integrating the boundary points and the historical trajectory, the structured map succinctly and effectively encapsulates the visited regions, thereby serving as input for the deep reinforcement learning based decision module. Instead of sequentially predicting the next action step within the decision module, we have implemented an advanced global planner to optimize the exploration path by leveraging long-range target points. We conducted experiments in three simulation environments and deployed in a real unmanned ground vehicle (UGV), the results demonstrate that our approach significantly reduces both the duration and distance of exploration compared with state-of-the-art methods.
\end{abstract}


\section{INTRODUCTION}
\IEEEPARstart{A}{ctive} SLAM represents a critical and intricate challenge within the field of robotics. While traditional SLAM approaches involve robots passively gathering sensory data followed by localization and map construction, Active SLAM underscores the autonomy of the robot. In this paradigm, the robot is tasked with actively selecting the most effective perceptual actions to improve the accuracy of positioning and the quality of mapping \cite{placed2023survey,garaffa2023reinforcement}. Essentially, the robot must exhibit three core competencies: constructing a map, ascertaining its position on this map, and adeptly managing its actions. The implications of Active SLAM extend broadly, with particular utility in search and rescue operations \cite{niroui2019deep}, as well as in exploration missions \cite{dang2020graph,azpurua2023survey}. In such contexts, robots are positioned to replace humans in venturing into challenging environments to conduct searches and surveys, and to autonomously create maps within uncharted territories.

The frontier-based exploration approach has emerged as a prevalent strategy for tackling the active SLAM challenge \cite{yamauchi1997frontier,ebadi2023present,wang2024exploration}. This method extracts regions at the boundaries between known obstacle-free space and unknown space within an occupancy grid map. In each iteration, the robot is directed towards the closest boundary point. While this method excels in decision-making efficiency, it requires the processing of the entire map to detect boundaries, resulting in a significant increase in computational resource consumption as the map size grows. Moreover, this approach does not account for path optimality. 

Sampling-based methods are known for their robustness against the scale of mapping environments \cite{umari2017autonomous,xu2021autonomous}. Among these, the Rapidly-Exploring Random Trees (RRT) method \cite{umari2017autonomous} stands out as a representative approach. While these methods exhibit resilience, their stochastic nature can lead to variable algorithmic performance, particularly in confined or intricately structured environments. This inconsistency often requires a substantial investment of time and a significant increase in the number of sampling points to identify feasible paths. Moreover, the paths they chart are frequently characterized by a higher degree of complexity and convolution.

Information-based methods leverage information theory to enhance robotic exploration strategies \cite{asgharivaskasi2023semantic,asgharivaskasi2022active,asgharivaskasi2025riemannian}. These approaches aim to maximize mutual information (MI) between the robot's actions and environmental map updates, thereby minimizing map entropy and reducing environmental uncertainty. Each exploration iteration prioritizes actions with the highest potential information gain. However, practical implementation faces challenges, including the intractability of optimal solutions and increased computational demands with larger exploration areas.

In recent years, DRL has gradually gained popularity. The adaptability of DRL renders it an auspicious candidate for tackling the Active SLAM problem \cite{niroui2019deep,chaplot2020learning,alcalde2022slam,chen2024lidar}. Unlike traditional methods that necessitate a pre-defined model of the environment, DRL enables robots to dynamically learn optimal policies through continuous interaction with their surroundings. This interactive learning process facilitates swift adaptation and autonomous decision-making within the crucible of unknown environments, equipping robots with the prowess to explore with efficacy. However, the current implementation of DRL in Active SLAM primarily revolves around teaching robots to execute single-step actions. This approach often overlooks the development of long-term planning strategies that can be sustained over extended periods. Consequently, robots may spend a disproportionate amount of time deliberating on decisions rather than executing actions, which can adversely impact exploration efficiency. Furthermore, there is a notable absence of strategies aimed at optimizing path length, a critical factor in the quest for more efficient exploration methodologies. In response to the aforementioned challenges, recent research has begun to focus on long-term decision-making processes. Some studies have integrated exploration strategies with path planning algorithms, enabling robots to output long-term executable target points rather than merely guiding single-step actions \cite{chaplot2020learning, chen2024lidar}. This integration allows robots to conduct more effective path planning and decision-making in complex environments, thereby enhancing exploration efficiency.

DRL-based active SLAM methods have demonstrated impressive performance in simple and small-scale environments. However, when it comes to large-scale and complex settings, many existing approaches are often hindered by inefficiencies, characterized by prolonged exploration times and suboptimal exploration paths. This study is dedicated to enhancing the efficiency and performance of active SLAM methods in large-scale complex indoor environments. To validate the performance of the proposed algorithms, we have developed both integrated simulation and real platforms that encompass perception, decision-making, planning, and control. The main contributions of this paper are as follows:
\begin{itemize}
\item We introduce a hierarchical and lightweight exploration framework that decouples high-level decision-making from low-level motion control,  enabling efficient large-scale exploration without requiring robots to learn fundamental navigation skills through trial-and-error.
\item We propose a novel structured map representation, combining discretized map partitioning, pose encoding, and dynamic boundary point processing.
\item We present a map-aware DRL with action refinement, which outputs intermediate waypoints rather than primitive actions and refined by an action optimization unit, improving valid decision.
\item We conduct comprehensive evaluation across $400-520 m^2$ simulated environments with distinct spatial challenges and evaluate our method within $20m \times 10m$ real world environment.
\end{itemize}

\section{RELATED WORK}

The frontier-based method introduced by \cite{yamauchi1997frontier} continues to be widely utilized in current research. Numerous scholars have sought to enhance this approach, aiming to reduce path lengths and computational costs \cite{orvsulic2019efficient}, \cite{sun2020frontier}. However, a comprehensive algorithm that effectively integrates these advantages has yet to be developed. In recent years, sampling-based methods have demonstrated promising performance, such as those utilizing RRT \cite{umari2017autonomous}, Rapidly-exploring Random Graphs \cite{dang2020graph}, and Probabilistic Roadmaps \cite{xu2021autonomous}. These methods primarily focus on evaluating the benefits of sampled paths, thereby circumventing the complexities associated with identifying and assessing all boundaries. Nevertheless, they are unstable and tend to degrade in scenarios where boundaries are sparse.

DRL has predominantly been instrumental in teaching robots single-step behaviors within small-scale simulation environments. For instance, Tai et al. \cite{tai2016robot} leveraged depth images as network inputs and employed the Deep Q-Network (DQN) algorithm to learn the robot's motion direction, successfully navigating it through corridors without colliding with walls. Zhang et al. \cite{zhang2017neural} introduced the Asynchronous Advantage Actor-Critic method, which facilitated robots in acquiring environmental representations and executing actions such as standing still, turning left or right, and moving forward. Their simulation tests confirmed the robot's ability to explore a room of approximately 23 square meters. In \cite{botteghi2020reinforcement}, three reward functions were designed based on the DQN algorithm, enabling robots to select appropriate actions in response to varying objectives, such as maximizing the mapped area or increasing map entropy. Expanding on this work, Botteghi et al. \cite{botteghi2021curiosity} introduced a curiosity-driven reward function to motivate robots to venture into uncharted territories. Their study involved training and testing in simulated environments spanning from 22 to 68 square meters. Zhao et al. \cite{zhao2024exploration} proposed and validated exploration-exploitation strategies in environments ranging from 22 to 135 square meters, confirming the method's efficacy. While DRL-based approaches for active SLAM have demonstrated promising autonomous decision-making capabilities, their reliance on single-step action selection limits efficiency in large-scale environments. Furthermore, existing studies have overlooked the critical challenge of optimizing for the shortest path during exploration, which remains a persistent gap in achieving robust and scalable solutions for complex navigation tasks.
 
 Recent advances in long-horizon robotic decision-making address exploration efficiency through hierarchical architectures. Chaplot et al.'s Active Neural SLAM (ANS) \cite{chaplot2020learning} employs a modular framework that decomposes global targets into executable sub-goals, improving 3D environment coverage. Chen et al. \cite{chen2024lidar} pioneer end-to-end reinforcement learning using RGB inputs instead of depth sensors, learning exploration policies from visual data. While these methods advance beyond geometric sensor dependence, two critical limitations persist: 1) high-dimensional action spaces in exploration tasks cause policies to converge to suboptimal local patterns, and 2) complex network architectures incur significant computational overhead, hindering real-time deployment on embedded systems.
 
 Furthermore, DRL often necessitates a substantial number of samples to train neural network models, which can lead to extended training durations and significant resource expenditure. To mitigate the curse of dimensionality and reduce computational resource usage, many DRL approaches opt to compress the observation space. For instance, Botteghi et al. \cite{botteghi2020reinforcement} introduced the concept of an observation window, where the parameter of observation window length is manually adjusted to determine how many past states are considered during each decision-making process. Chen et al. \cite{chen2023multi} performed uniform sampling on laser data with a detection angle of 270$^\circ$, thereby reducing the observation space data volume from 1080 to 24. Similarly, Alcalde et al. \cite{alcalde2022slam} utilized only five lidar data points, uniformly distributed within the robot's 180$^\circ$ frontal view, as inputs for the network. However, such compression of the observation space can diminish the available information, potentially leading to the loss of critical details and adversely impacting the agent's environmental comprehension and decision-making capabilities. We propose a structured map encoding framework for DRL-based exploration, compressing environmental topology into spatial-semantic tensors. Unlike methods using raw observations, our structured representation enables direct network input of hierarchical features, improving computational efficiency while preserving navigational constraints critical for learning coherent exploration strategies.

\section{APPROACH}
\subsection{Problem Statement}
	Active SLAM optimizes robotic exploration in unknown 2D environments by sequential decision-making under uncertainty, formalized as a Partially Observable Markov Decision Process (POMDP) with components $(\mathcal{S},\mathcal{A},\mathcal{T},\mathcal{R},\mathcal{O},\gamma)$:
	state space $\mathcal{S}$, action space $\mathcal{A}$, transition model $\mathcal{T}$, reward function $\mathcal{R}$, observation space $\mathcal{O}$, and discount factor $\gamma$. At each step, the agent executes action $a_t$ under policy $\pi$, transitions to $s_{t+1} \sim \mathcal{T}(\cdot|s_t,a_t)$ and receives observation $o_{t+1}$. The objective of the agent is to find the optimal strategy $\pi*$ that maximizes the expected return of each state-action pair, as shown in Eq. \ref{eq1}.
\begin{equation}\label{eq1}
	\pi^* = \arg\max_{\pi} \, \mathbb{E} \left[ \sum_{t=0}^{\infty} \gamma^t \mathcal{R}(s_t, a_t) \right]
\end{equation}

\subsection{System Overview}
\begin{figure*}[htb]
	\centering         
	\includegraphics[width=0.9\textwidth]{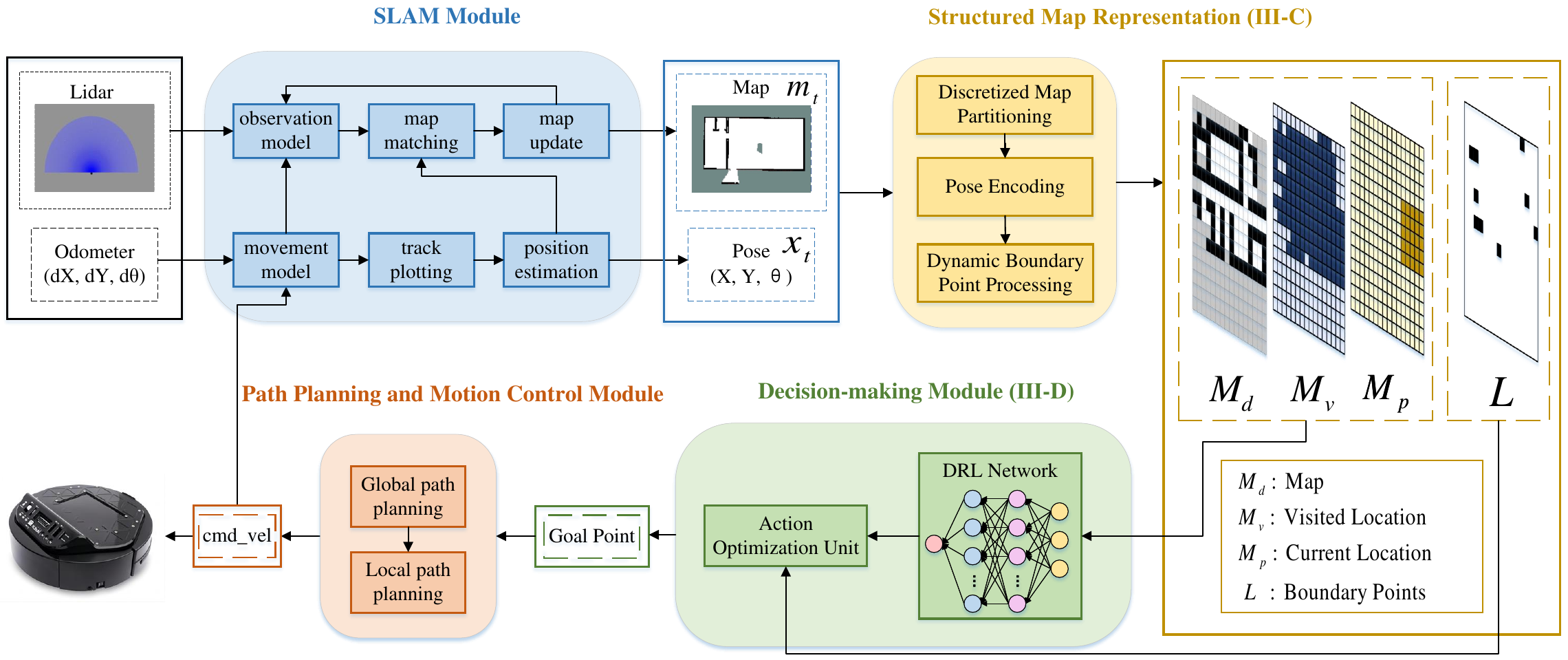} 
	\caption{\small The overall framework of the robot autonomous exploration system.}
	\label{fig:overview}
\end{figure*}

The proposed autonomous exploration system is composed of SLAM module, structured map representation, decision-making module and path planning and motion control module, as shown in Fig. \ref{fig:overview}.
The SLAM module is responsible for self-localization and incremental mapping of the surrounding environment by gathering Lidar and odometer data. The SLAM problem can be formulated as follows:

\begin{equation}
	\begin{gathered}
	{x_t} = f({x_{t - 1}},{u_t}) + {\delta _{act}},\\
	{m_t} = g({m_{t - 1}},{z_t},{x_t}) + {\delta _{sen}},
	\end{gathered}
\end{equation}
where $f$ and $g$ denote the motion and measurement model, $\delta _{act}$ signifies actuation noise and $\delta _{sen}$ indicates sensor noise. The state of the robot at time $t$ is denoted as ${x_t}$, which is related to its state at the previous time step $x_{t - 1}$ and the action taken ${u_t}$. Concurrently, the map constructed by the robot at time $t$, denoted as ${m_t}$, is associated with the map from the previous time step ${m_{t-1}}$ and the observation ${z_t}$.

This work employs the Gmapping algorithm \cite{grisetti2007improved} as the SLAM framework, prioritizing its computational efficiency and high-fidelity mapping capabilities. The proposed structured map representation processes pose-graph and occupancy grid outputs from SLAM to extract navigation-critical features, including spatial relationships between explored and frontier regions and topological connectivity of traversable areas. As detailed in Section \ref{sec:B}, these features are encoded into a compact three-channel tensor ($3\times F\times G$) that preserves environmental semantics. This structured representation feeds into the Map-Aware DRL, where a DRL agent iteratively selects the optimal navigation waypoint for the next planning horizon. The policy network’s architecture and training paradigm, which leverage partial observability constraints, are formalized in Section \ref{sec:C}.
Upon receiving target coordinates from the decision module, our hierarchical motion planner generates collision-free trajectories using the Robot Operating System (ROS) \cite{quigley2009ros} navigation stack. The global planner employs an A* variant \cite{hart1968formal} for optimal path finding in known environments, while the TEB algorithm \cite{rosmann2012trajectory} dynamically adjusts velocity profiles for real-time obstacle avoidance.

\subsection{Structured Map Representation}
\label{sec:B}
\begin{figure*}
	\setlength{\abovecaptionskip}{0.1cm}
	\centering
	\includegraphics[width=0.9\textwidth]{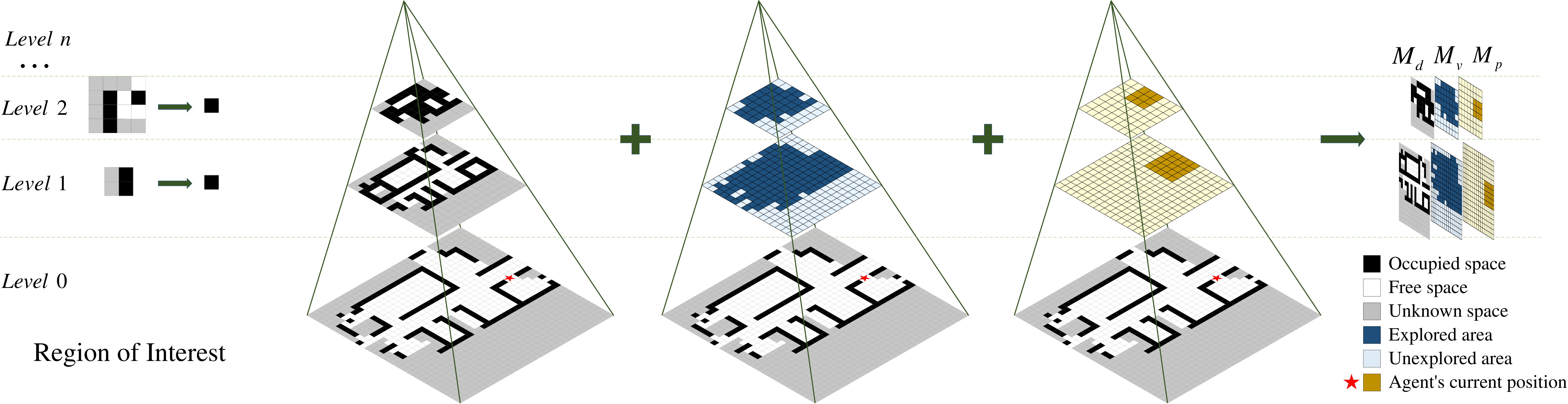}	
	\caption{\small Example of Discretized Map Partitioning and Pose Encoding.}
	\label{fig:Example of Map Structuring and Pose Structuring}
\end{figure*}

The real-time processing of a substantial amount of data in exploring large-scale complex environments is challenging since the computational resources are limited. Therefore, we proposed an innovative structured map representation, as illustrate in Fig. \ref{fig:Example of Map Structuring and Pose Structuring}, to facilitate rapid analysis and efficient management of map and pose data from the SLAM module, expressed as $M=S({m_t},{x_t})$.

\subsubsection{Discretized Map Partitioning}
To enable systematic environment exploration, we partition the occupancy grid into $n \times n$ regions of interest (ROIs). Each ROI is classified through our novel semantic mapping $M_d(x, y)$, which evaluates spatial awareness requirements using:

\begin{equation}
{M_d}(x,y) = 
	\begin{cases}
		1, & \text{if } C_o \ge \alpha n^2 \\
		0, & \text{if } C_o < \alpha n^2 \text{ and } C_f \ge \beta C_u , \\
		-1, & \text{otherwise}
	\end{cases}
\end{equation}

where ${M_d}$ is the post-assignment map, $C_o$, $C_f$, and $C_u$ represent the counts of occupied, free, and unknown cells, respectively. Increasing $\beta$ raises the free-to-unknown ratio requirement for navigable region designation, thereby prioritizing exploration to underobserved areas. Conversely, elevating $\alpha$ tightens the obstacle density threshold for exploration-priority regions, systematically heightening risk aversion in cluttered environments. 

At the same time, the boundary point region is marked to delineate an area devoid of occupied grids yet encompassing both free and unknown grids. Finally, a new binary map  ${M_v}$ is created with the same dimensions as the map  ${M_d}$, and is utilized to meticulously document all regions that have been visited:
\begin{equation}
{M_v}(x,y) = 1 \quad if \ {M_d}(x,y) \ne  - 1,
\end{equation}
where ${M_v}$ denotes the area that has been explored.

\subsubsection{Pose Encoding}
We generate a real-time egocentric spatial mask $M_p \in \{0, 1\}^{F \times G}$ isomorphic to the base map $M_d$, where:
\begin{equation}
	M_p(x, y) = \begin{cases}
		1, & \text{if } \|(x, y) - (x_r, x_y)\|_\infty \leq \lambda \\
		0, & \text{otherwise}
	\end{cases},
\end{equation}
Here, $(x_r, y_r)$ denote the robot's grid-aligned coordinates, and $\| \cdot \|_\infty \leq \lambda$ defines a Chebyshev distance neighborhood of radius $\lambda$.

\subsubsection{Dynamic Boundary Point Processing} 
The initial list of boundary points is further processed as follows.
Firstly, in order to facilitate subsequent navigation, the list of boundary point regions is converted to coordinates in the map coordinate system:
\begin{equation}
\begin{aligned}
x &= {x_{origin}} +n \times {i_{width}} \times r,\\
y &= {y_{origin}} +n \times {j_{height}} \times r,
\end{aligned}
\end{equation}
where ($x_{origin},y_{origin}$) are the top-left corner of the map, $i_{width}$ and $i_{height}$ are the index of the boundary points in the grid, $r$ is the resolution of the occupied grid map and $n$ is the size of the ROI. Then we obtained the set of the boundary points, denoted as:
$L' = \{ ({x_0}^\prime ,{y_0}^\prime ),({x_1}^\prime ,{y_1}^\prime ), \ldots ,({x_n}^\prime ,{y_n}^\prime )\}$.

We perform density-based feasibility screening on boundary points $l \in L'$ to identify navigable regions. For each candidate $l$, we evaluate its $k$-neighborhood (radius R) using:
\begin{equation}
	\mathcal{N}(l,R) = \{ {({x_i},{y_i})| {\sqrt {{{({x_i} - {x_l})}^2} + {{({y_i} - {y_l})}^2}}  \le R} .} \}.
\end{equation}

A point $l$ is deemed feasible if $\mathcal{N}(l,R)\ge k$, where $k$ is the density threshold ensuring spatial continuity. To incorporate operational constraints, we maintain a dynamically updated exclusion set $Q$ containing boundary points proven unreachable due to obstacles or topological  disconnects (e.g., dead-ends). The feasible set $L_b$ is thus:

\begin{equation}
	{L_b} = \left\{ {l\left| {\mathcal{N}(l,R)} \right. \ge k,l \in L',l\not\in Q} \right\}.
\end{equation}

We apply MeanShift clustering with adaptive bandwidth estimation to compress boundary points while preserving spatial distribution characteristics:
\begin{equation} 
{L_c} = Meanshift({L_b}).
\end{equation}

\begin{figure}[htbp]
	\centering
	\begin{subfigure}[b]{0.235\textwidth}
		\centering
		\includegraphics[width=\textwidth]{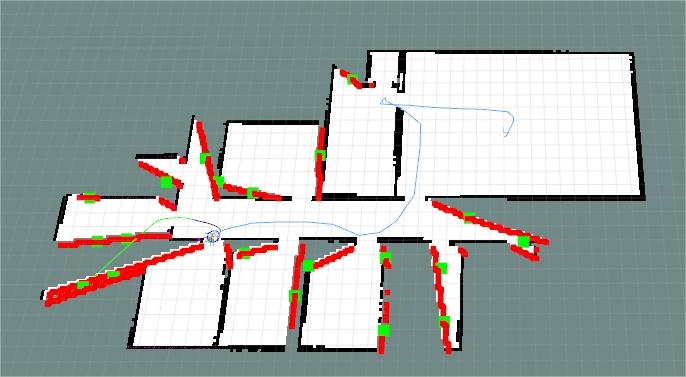}
		\caption{\small}
		\label{fig:1}
	\end{subfigure}
	\begin{subfigure}[b]{0.235\textwidth}
		\centering
		\includegraphics[width=\textwidth]{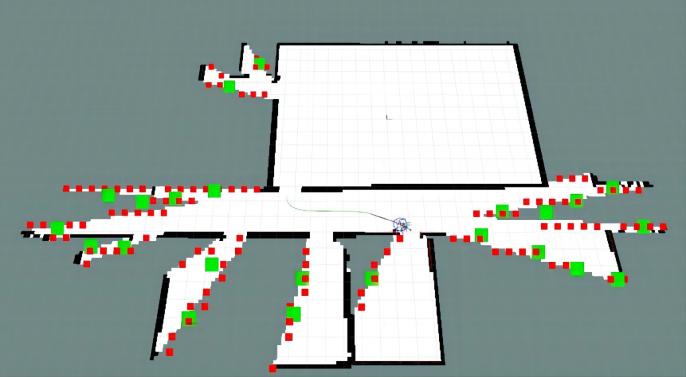}
		\caption{\small}
		\label{fig:2}
	\end{subfigure}
	\caption{
		\small Boundary Point Detection Example. Red dots are boundary points; green dots are cluster centers. (a) Before processing; (b) after structured map representation.}
	\label{fig:detected boundary points examples}
\end{figure}

Each cluster centroid $u \in {L_c}$ is then mapped to its nearest navigable proxy:
\begin{equation} 
L = \bigcup\limits_{u \in {L_c}} {\mathop {\arg \min }\limits_{m \in {L_b}} d(u,m)},
\end{equation}
Fig. \ref{fig:detected boundary points examples} illustrates an example of boundary point detection. 

After the above processing, the final information passed to the decision module includes a list of representative boundary points $L$ and a dimensionality reduction map of three channels, denoted as $M' = cat\{ {M_d};{M_v};{M_p}\}$.

\subsection{Map-Aware DRL}
\label{sec:C}
Our implementation builds upon the Proximal Policy Optimization (PPO) framework [28] with two critical enhancements for exploration-constrained navigation:
\begin{equation}\label{eq:reward}
	r_t = \underbrace{\mu \Delta\mathcal{C}_t}_{\text{Coverage}} 
	- \underbrace{\|L_t - L_{t-1}\|_2}_{\text{Path Smoothness}},
\end{equation}
where $\Delta\mathcal{C}_t$ is the new coverage area, $\|L_t - L_{t-1}\|$ penalizes erratic distance changes.
\subsubsection{Action Optimization Unit (AOU)}
\begin{figure}[htb]
	\centering         
	\includegraphics[width=1.0\linewidth]{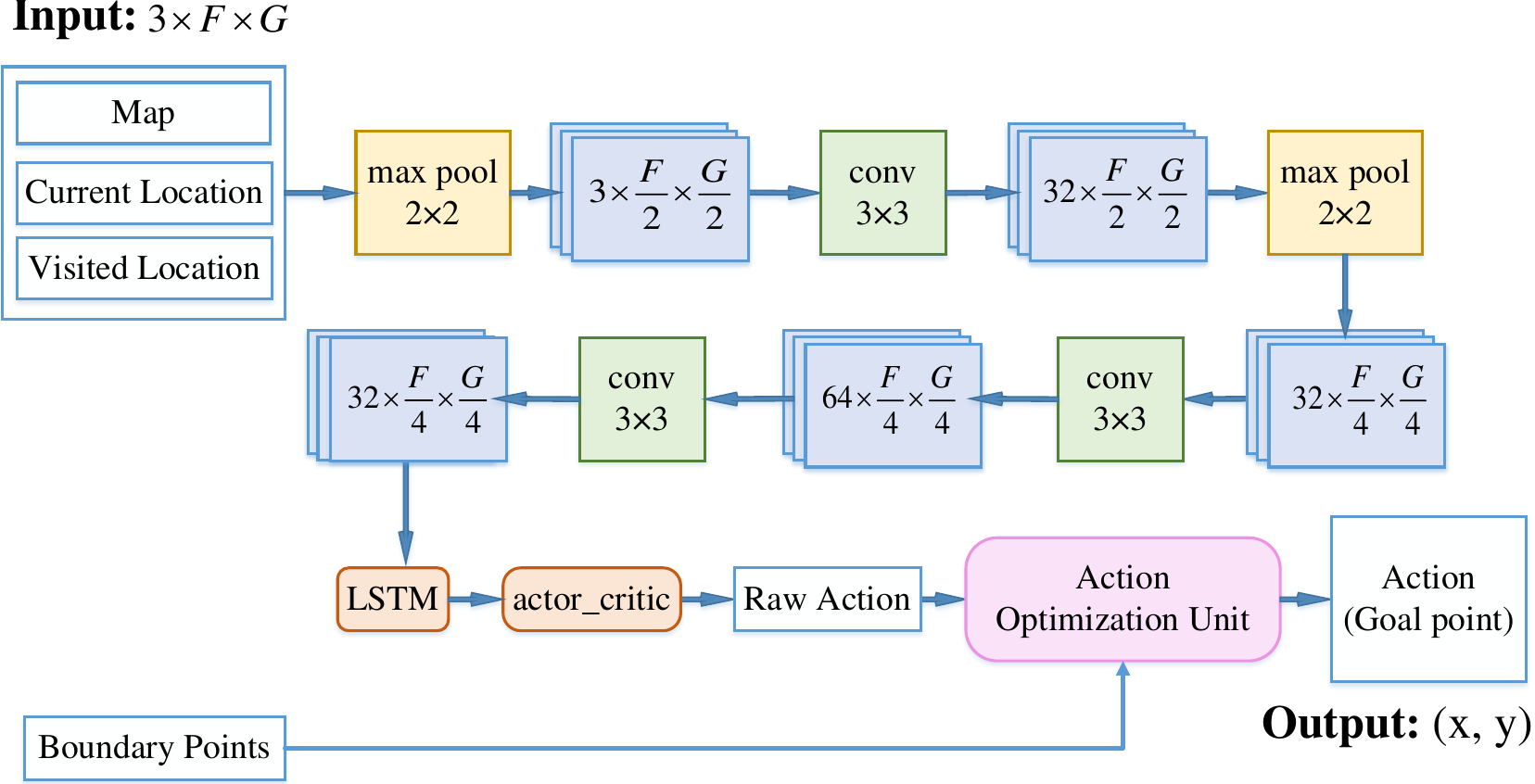} 
	\caption{Network structure of the decision module.}
	\label{fig:network}
\end{figure}

\begin{figure}[htb]
	\centering         
	\includegraphics[width=0.7\linewidth]{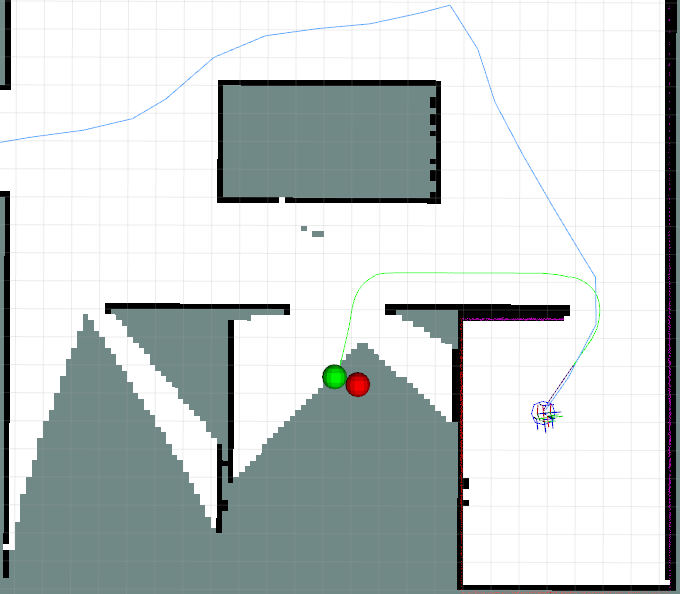} 
	\caption{Schematic Diagram of Action Optimization Unit Application.}
	\label{fig:AOU_sample}
\end{figure}

As depicted in Fig. \ref{fig:network}, our decision network processes a 3-channel occupancy grid tensor $M \in \mathbb{R}^{3 \times F \times G}$ (occupied/free/unknown) through a hybrid architecture combining a spatial encoder and a 32-unit LSTM \cite{hochreiter1997long}. The network outputs normalized coordinates $(x',y')\in [0,1]^2$. A critical challenge arises when the initial network prediction (marked in red in Fig. \ref{fig:AOU_sample}) falls within unknown regions, where no valid exploration path exists due to the absence of prior maps. To address this, the Action Optimization Unit projects these coordinates onto the nearest information-rich boundary point $l^*\in L$ (marked in green in Fig. \ref{fig:AOU_sample}). This selection prioritizes boundary points, located at the interface of known and unknown regions, which are rich in exploration potential due to their high information value. By leveraging these strategic points, the AOU ensures the robot prioritizes areas critical for expanding environmental understanding, enabling precise and adaptive navigation in complex scenarios. This optimization mechanism bridges high-level decision-making with actionable navigation targets, improving both computational efficiency and exploration effectiveness.

\subsubsection{Observation-Action-State}
The observation space $\mathcal{O}$ comprises a three-channel map: the first channel encodes the current map state, the second marks visited locations, and the third indicates the robot’s relative position. The action space $\mathcal{A}$ is a 2D vector of floating-point values ($[-1.0, 1.0]$), sampled from the map and convertible to real-world coordinates based on map resolution. The state space $\mathcal{S}$ at time $t$, defined as $\mathcal{S}_t = (\mathcal{O}_t, a_{t-1}, \mathcal{A}_{t-1})$, integrates the current observation $\mathcal{O}_t$ (encompassing pose, map data, and visited areas), the prior action $a_{t-1}$, and the post-action map completeness $\mathcal{A}_{t-1}$ to inform the next decision. This structure links environmental perception, historical actions, and iterative map updates to guide navigation.
\section{EXPERIMENTS AND RESULTS}
\subsection{Experimental Settings}
Our exploration strategy is trained using the Gazebo simulation platform, a high-fidelity environment designed to accurately replicate real-world physical robot interactions.

\begin{figure}[htb]
	\centering         
	\includegraphics[width=\linewidth]{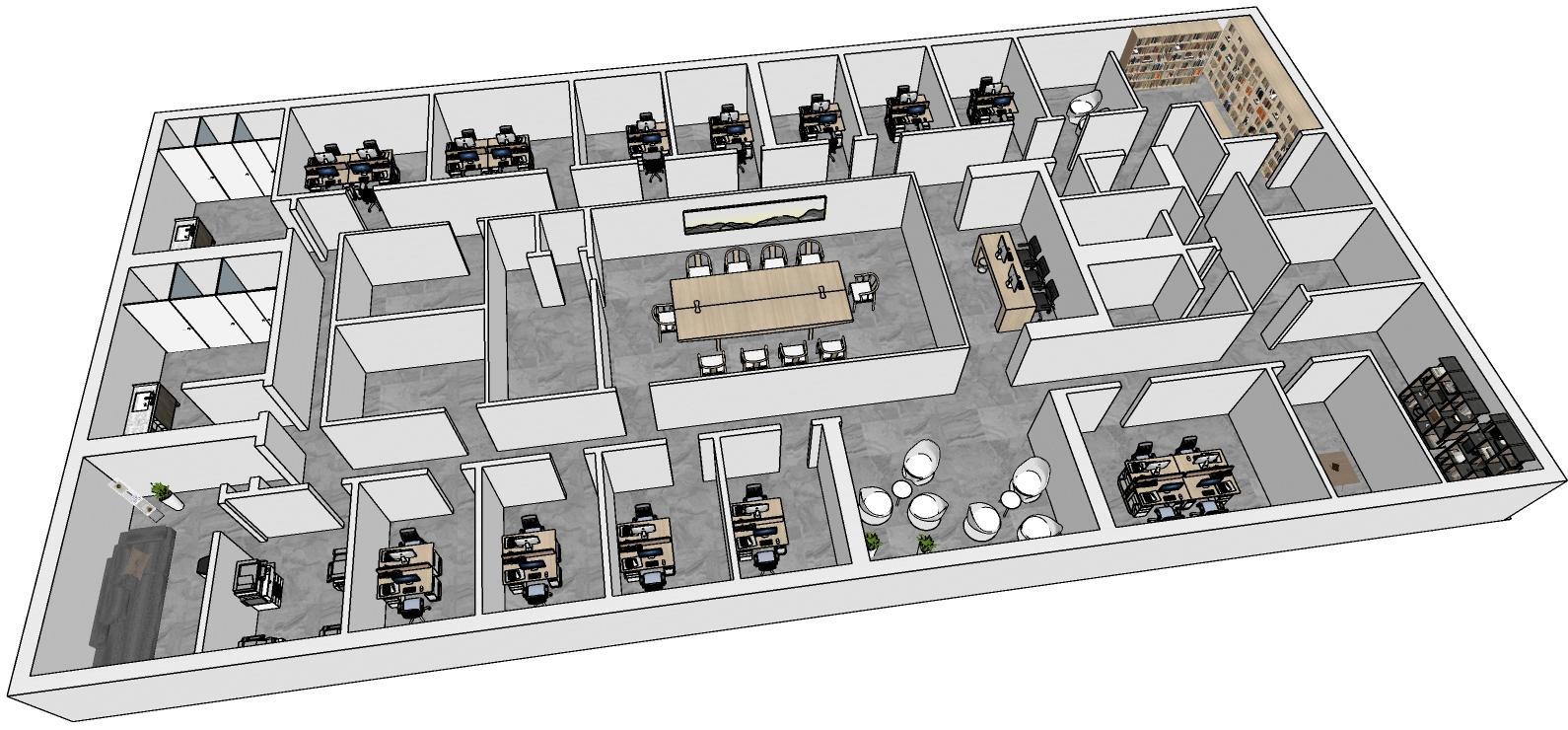} 
	\caption{\small Env-1 for training.}
	\label{fig:env1}
\end{figure}

The robotic model is configured with a 180° horizontal field of view and an 8-meter measurement range, with all sensor noise in simulations modeled using a Gaussian distribution. Critical training hyperparameters are summarized in Table \ref{param}. As illustrated in Fig. \ref{fig:env1}, the primary training environment (Env-1) replicates a 520 $m^2$ indoor space featuring complex room layouts and obstacle arrangements to mirror real-world operational challenges. To rigorously evaluate adaptability, the agent is tested across three additional environments (Env-2 to Env-4), spanning 400–520 $m^2$ with distinct spatial configurations: Env-2 emphasizes navigation through elongated corridors, Env-3 incorporates a high obstacle density and intricate pathways, and Env-4 simulates clustered small-room scenarios to stress-test spatial reasoning.

\begin{table}[h]
	\caption{Training and simulation main hyperparameters.}
	\centering
	\begin{tabular}{cccc}
		\toprule  
		Hyperparameters & value \\
		\midrule 
		size of the ROI $n$ & 4 \\
		Weighting factors $\alpha, \beta$ & 0.18, 1.0 \\
		Uncertainty modifier $\lambda$ & 3 \\
		Map resolution $r$ & 0.1 \\
		Map size $F,G$ & 32, 64 \\
		Feasibility Radius $R$ & 1 \\
		Minimum Neighbor Count $k$ & 3 \\
		Scaling factors $\mu$ & 0.02 \\
		Batch size & 32 \\
		Episode length & 40 \\
		Discount factor  & 0.99 \\
		Learning rate & 0.00025 \\
		\bottomrule  
	\end{tabular}
	\label{param}
\end{table}

\begin{table*}
	\centering
	\caption{Evaluation results in Env-2, Env-3, Env-4.}
	\resizebox{\linewidth}{!}{
	\begin{tabular}{cccccccccc} 
		\toprule
		\multirow{2}*{Env} & \multirow{2}*{Method} & \multicolumn{2}{c}{Mostly Completed($\geq$75\%)} & \multicolumn{2}{c}{Essentially Completed($\geq$93\%)} & \multicolumn{3}{c}{Naturally Stopped} & \multirow{2}*{Test Failures} \\
		
		\cmidrule(lr){3-4}
		\cmidrule(lr){5-6}
		\cmidrule(lr){7-9}
		
		& & Time(s) & Length(m) & Time(s) & Length(m) & Map completeness(\%) & Time(s) &Length(m)  & \\
		
		\midrule 
		\multirow{3}*{Env-2} & Ours & \textbf{217.23} & \textbf{81.69} & \textbf{322.93} & 124.67 & 99.98& \textbf{356.45} & \textbf{136.75} & {No}\\ 
		& RRT & 283.47 & 91.03 & 366.17 & \textbf{122.37} & \textbf{100} & 423.70 & 143.37 & {Yes}\\
		& Frontier & 260.78 & 96.09 & 333.87 & 122.88 & \textbf{100} & 385.24 & 144.94 & {No}\\
		& TARE & 274.28 & 108.67 & 377.63 & 144.80 & 99.92 & 437.28 & 167.25 & {Yes}\\
		& DA-SLAM & - & - & - & - & 51.93 & 303.88 & 128.77 & {Yes}\\
		
		\midrule 
		\multirow{3}*{Env-3} & Ours & \textbf{287.25} & \textbf{116.31} & \textbf{365.98} & \textbf{146.95} & 99.05 & \textbf{407.68} & \textbf{164.68} & {No}\\ 
		& RRT & 315.02 & 125.31 & 465.28 & 191.00 & 99.04 & 528.41 & 215.73 & {No}\\
		& Frontier & 328.76 & 122.78 & 453.82 & 171.45 & 99.09 & 465.85 & 175.38 & {No}\\
		& TARE & 308.96 & 123.54 & 372.93 & 150.21 & \textbf{99.38} & 472.41 & 191.05 & {Yes}\\
		& DA-SLAM & - & - & - & - & 35.90 & 561.25 & 238.25 & {Yes}\\
		
		\midrule 
		\multirow{3}*{Env-4} & Ours & \textbf{346.06} & 128.63 & \textbf{532.76} & \textbf{208.52} & 97.96 & \textbf{612.58} & 236.51 & {No}\\ 
		& RRT & 366.36 & 129.18 & 632.19 & 228.15 & \textbf{98.98} & 847.30 & 307.28 & {Yes}\\
		& Frontier & 390.69 & 136.81 & 586.66 & 210.19 & 98.02 & 662.89 & 240.20 & {No}\\
		& TARE & 348.01 & \textbf{115.60} & - & - & 90.12 & 621.07 & \textbf{204.09} & {Yes}\\
		& DA-SLAM & - & - & - & - & 34.24 & 554.36 & 231.19 & {Yes}\\
		\bottomrule 
		\label{table:result}
	\end{tabular}
}
\end{table*}

\begin{figure*}
	\centering
	\begin{subfigure}[b]{0.16\textwidth}
		\includegraphics[width=\textwidth]{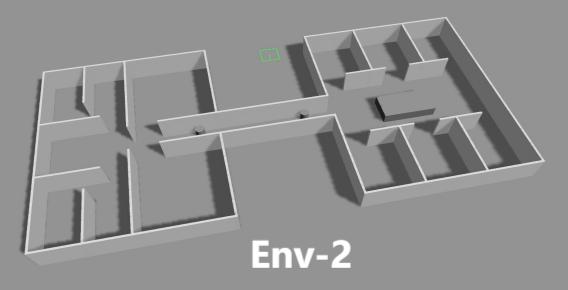}
	\end{subfigure}
	\begin{subfigure}[b]{0.16\textwidth}
		\includegraphics[width=\textwidth]{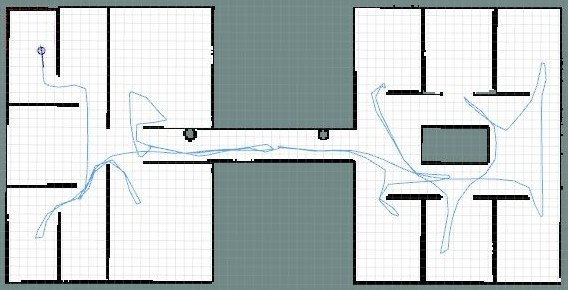}
	\end{subfigure}
	\begin{subfigure}[b]{0.16\textwidth}
		\includegraphics[width=\textwidth]{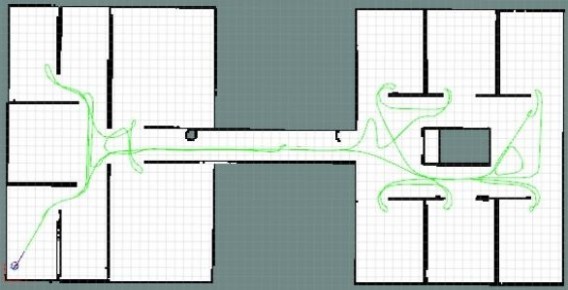}
	\end{subfigure}
	\begin{subfigure}[b]{0.16\textwidth}
		\includegraphics[width=\textwidth]{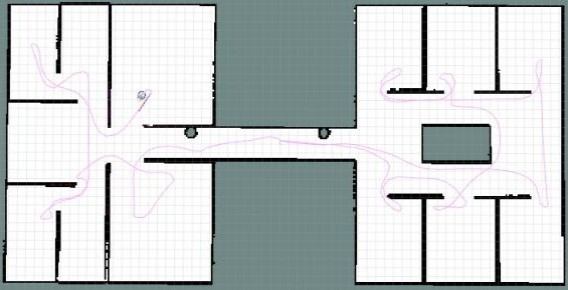}
	\end{subfigure}
	\begin{subfigure}[b]{0.16\textwidth}
		\includegraphics[width=\textwidth]{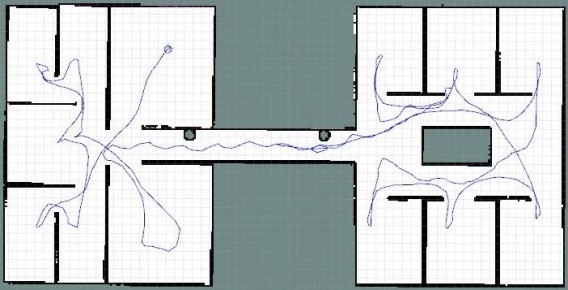}
	\end{subfigure}
	\begin{subfigure}[b]{0.16\textwidth}
		\includegraphics[width=\textwidth]{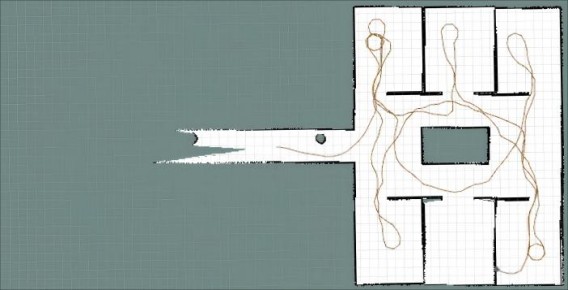}
	\end{subfigure}	
	\vspace{0.1cm}
	
	\begin{subfigure}[b]{0.16\textwidth}
		\includegraphics[width=\textwidth]{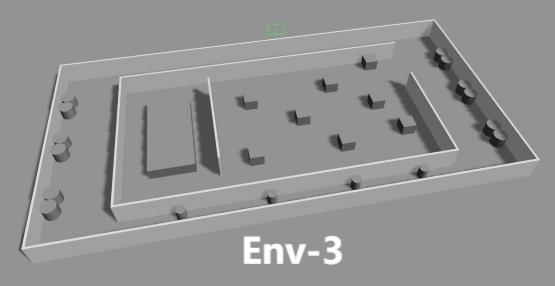}
	\end{subfigure}	
	\begin{subfigure}[b]{0.16\textwidth}
		\includegraphics[width=\textwidth]{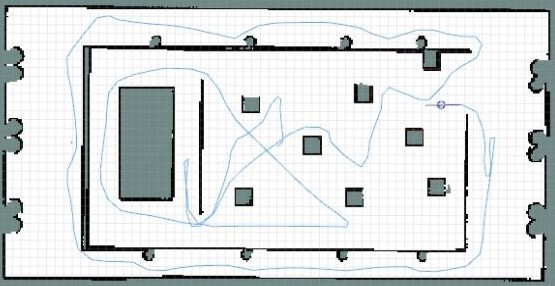}
	\end{subfigure}
	\begin{subfigure}[b]{0.16\textwidth}
		\includegraphics[width=\textwidth]{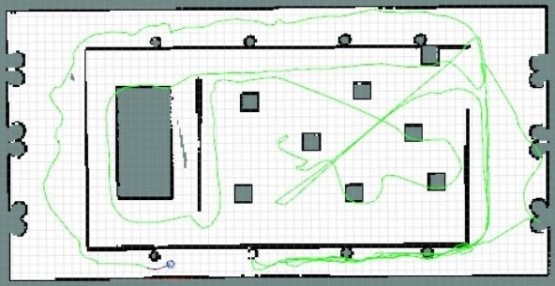}
	\end{subfigure}
	\begin{subfigure}[b]{0.16\textwidth}
		\includegraphics[width=\textwidth]{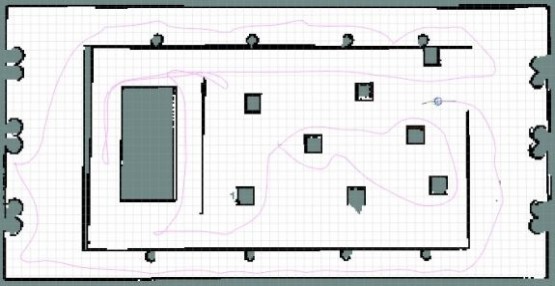}
	\end{subfigure}
	\begin{subfigure}[b]{0.16\textwidth}
		\includegraphics[width=\textwidth]{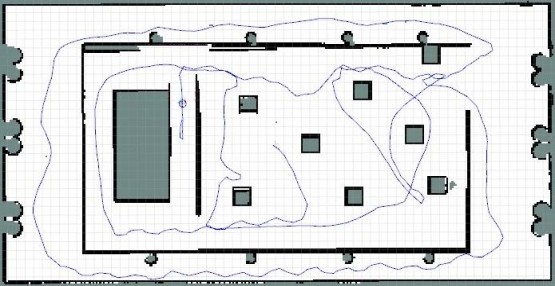}
	\end{subfigure}
	\begin{subfigure}[b]{0.16\textwidth}
		\includegraphics[width=\textwidth]{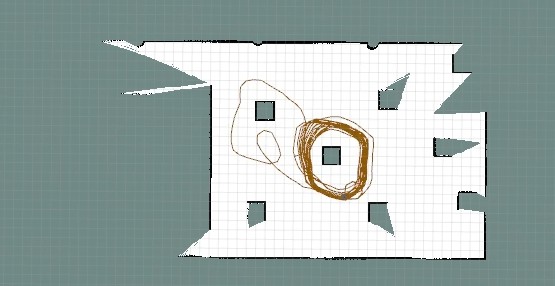}
	\end{subfigure}
	\vspace{0.1cm}
	
	\begin{subfigure}[b]{0.16\textwidth}
		\includegraphics[width=\textwidth]{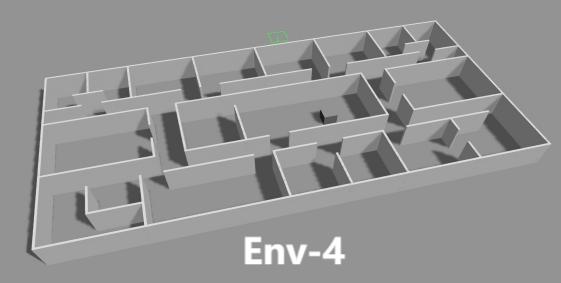}
		\caption{\small Env}
	\end{subfigure}
	\begin{subfigure}[b]{0.16\textwidth}
		\includegraphics[width=\textwidth]{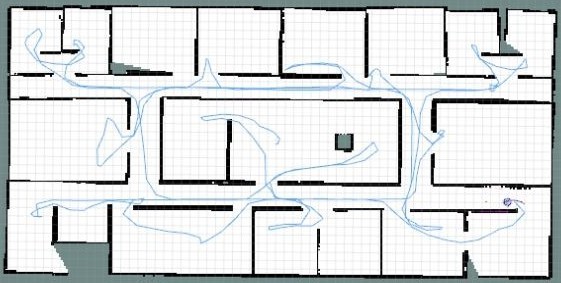}
		\caption{\small Ours}
	\end{subfigure}
	\begin{subfigure}[b]{0.16\textwidth}
		\includegraphics[width=\textwidth]{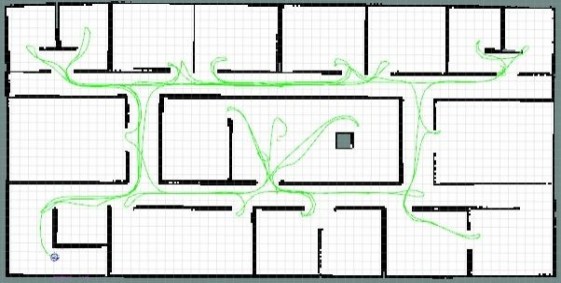}
		\caption{\small RRT}
	\end{subfigure}
	\begin{subfigure}[b]{0.16\textwidth}
		\includegraphics[width=\textwidth]{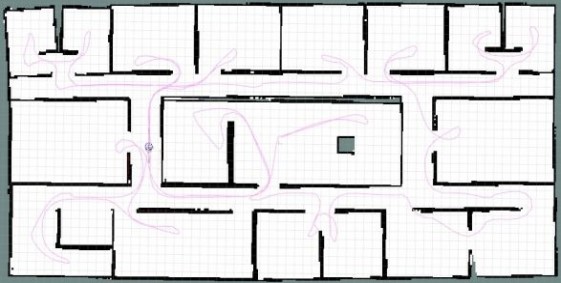}
		\caption{\small Frontier}
	\end{subfigure}
	\begin{subfigure}[b]{0.16\textwidth}
		\includegraphics[width=\textwidth]{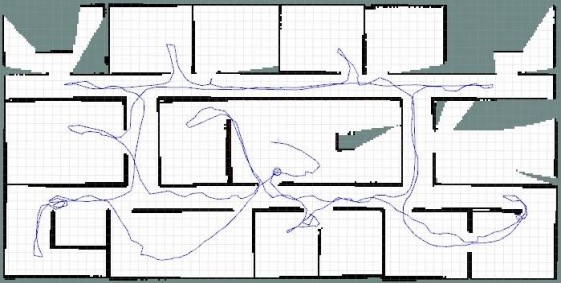}
		\caption{\small TARE}
	\end{subfigure}
	\begin{subfigure}[b]{0.16\textwidth}
		\includegraphics[width=\textwidth]{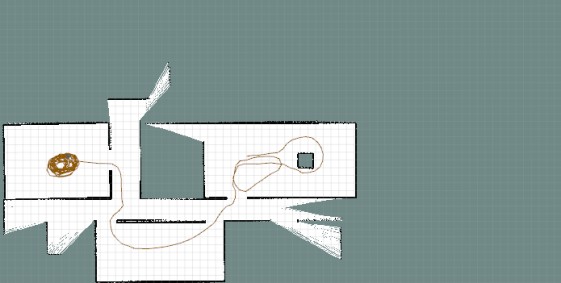}
		\caption{\small DA-SLAM}
	\end{subfigure}
	\vspace{0.1cm}
	
	\caption{Trajectories and mapping results of Env-2, Env-3, Env-4 for Ours, RRT, Frontier, TARE, DA-SLAM.}
	\label{fig:result}
\end{figure*}

\begin{figure}
	\centering
	\begin{subfigure}[b]{0.235\textwidth}
		\centering
		\includegraphics[width=\textwidth]{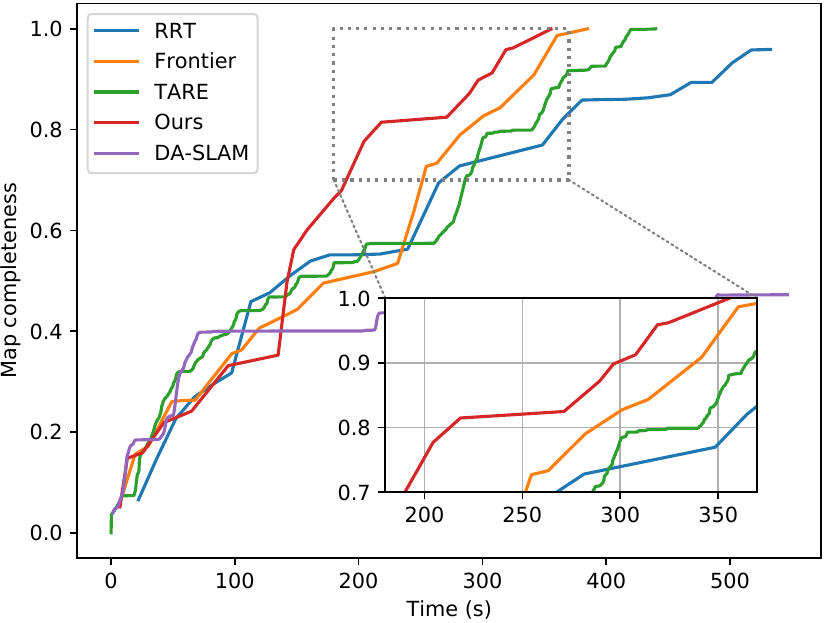}
		\label{fig:1}
	\end{subfigure}
	\hfill
	\begin{subfigure}[b]{0.235\textwidth}
		\centering
		\includegraphics[width=\textwidth]{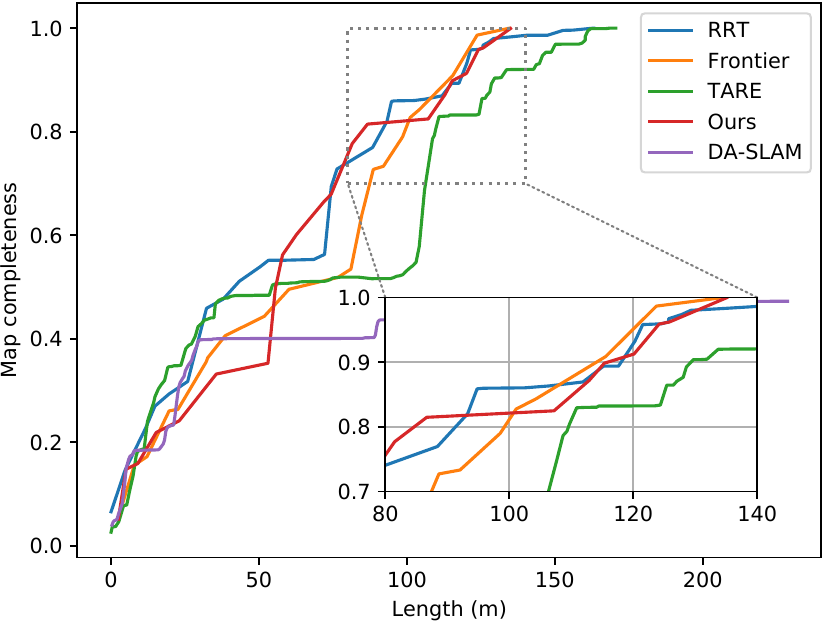}
		\label{fig:2}
	\end{subfigure}
	\vspace{-0.5cm}
	\caption*{\small(a) test Env-2}
	\vspace{0.2cm}
	
	\begin{subfigure}[b]{0.235\textwidth}
		\centering
		\includegraphics[width=\textwidth]{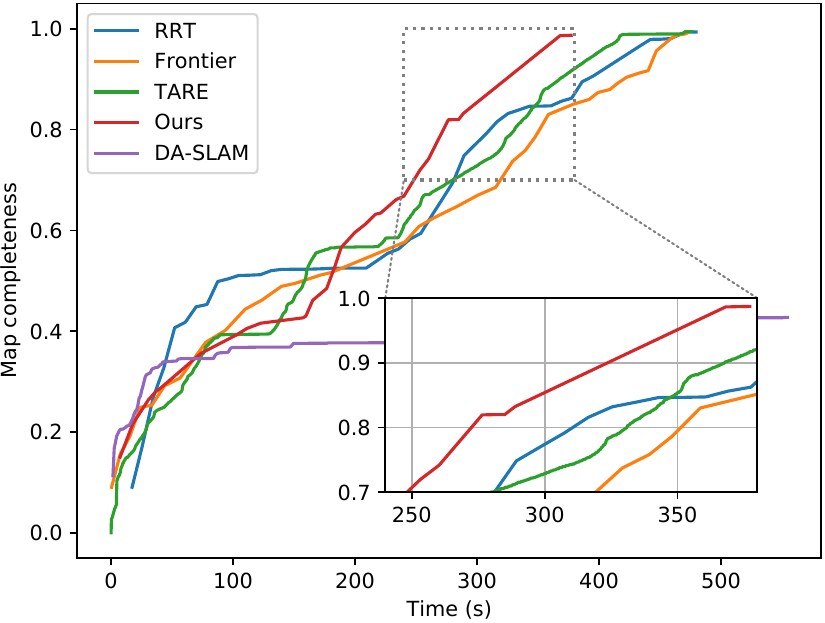}
		\label{fig:3}
	\end{subfigure}
	\hfill
	\begin{subfigure}[b]{0.235\textwidth}
		\centering
		\includegraphics[width=\textwidth]{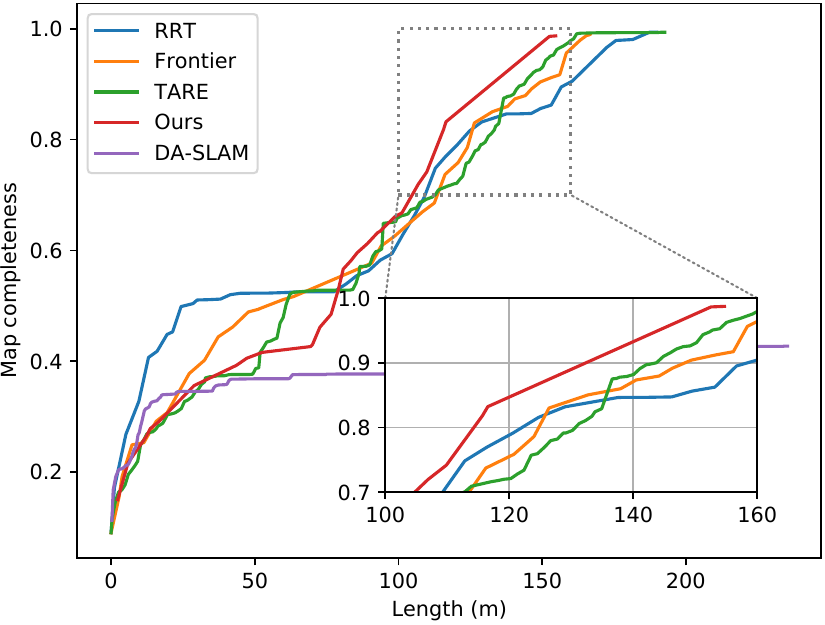}
		\label{fig:4}
	\end{subfigure}
	\vspace{-0.5cm}
	\caption*{\small(b) test Env-3}
	\vspace{0.2cm}
		
	\begin{subfigure}[b]{0.235\textwidth}
		\centering
		\includegraphics[width=\textwidth]{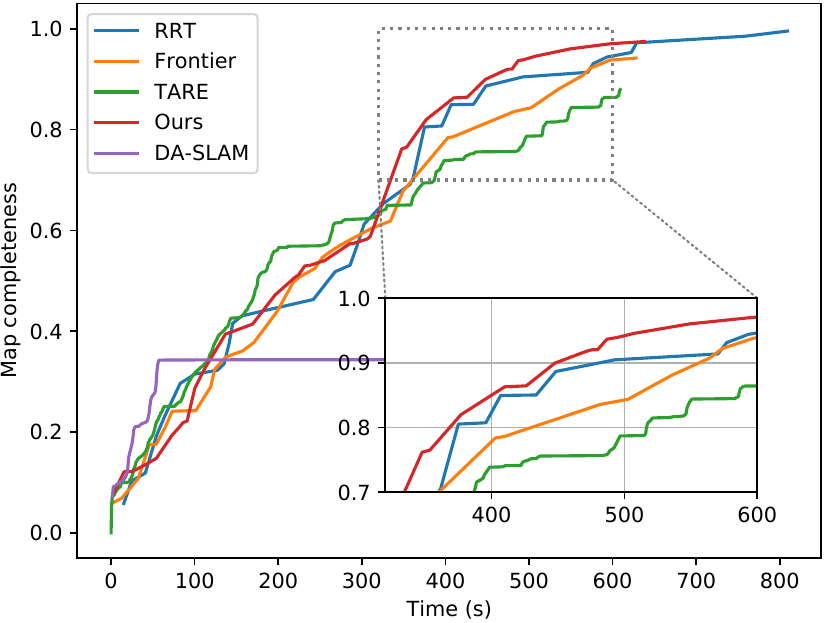}
		\label{fig:5}
	\end{subfigure}
	\hfill
	\begin{subfigure}[b]{0.235\textwidth}
		\centering
		\includegraphics[width=\textwidth]{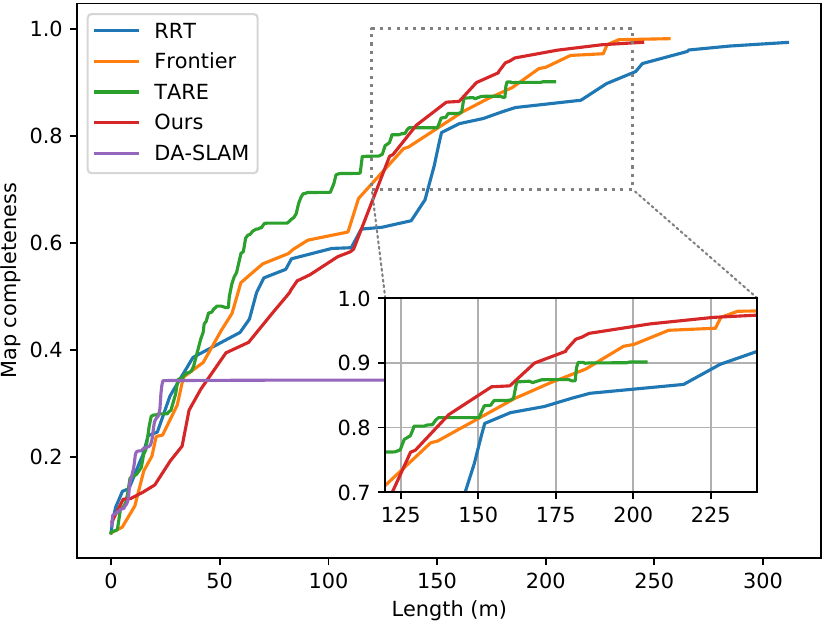}
		\label{fig:6}
	\end{subfigure}
	\vspace{-0.5cm}
	\caption*{\small(c) test Env-4}
	\vspace{0.3cm}
	\caption{Algorithm exploration progress in three environments over time and path length. Left: map completion rate over time; Right: map completion rate over path length.}
	\label{fig:images}
\end{figure}

\subsection{Training and Testing Evaluation}
The training protocol structures learning through episodic iterations, where each episode represents a distinct exploration task. In Env-1, episodes end either upon successful completion of the exploration or once a 40-step limit is reached. Training concludes when per-episode cumulative rewards stabilize and asymptotically approach peak performance levels, indicating policy convergence. The validated model is then assessed in unseen testing scenarios to evaluate generalization across diverse environmental challenges.

Cross-environment benchmarking against established methods, frontier-based exploration \cite{yamauchi1997frontier}, RRT-based planning \cite{umari2017autonomous}, TARE \cite{cao2021tare}, and DA-SLAM \cite{alcalde2022slam}, quantifies the empirical performance of our strategy and evaluates its adaptability across diverse environmental layouts.

We quantify exploration performance using two metrics: total time and total path length. Exploration completion is classified into three tiers: mostly completed, essentially completed, and naturally stopped. Mostly completed refers to when the explored area reaches 75\% or more. Considering potential mapping errors, essentially completed is defined as the explored area reaching 93\% or more \cite{botteghi2021curiosity}. 
Natural termination varies across algorithms due to differing target-selection methodologies. For instance, frontier-based methods terminate when no new frontiers remain, while sampling-based approaches (e.g., RRT) halt upon coverage saturation. To ensure equitable comparison, we recorded both the time-to-termination and map completion percentage at each algorithm’s natural stopping point. This dual-metric approach enables systematic assessment of efficiency (time/path) and coverage robustness across heterogeneous strategies.

In each environment, we conducted tests for each algorithm more than ten times, the average results are shown in Table \ref{table:result}. The final exploration trajectories and mapping results for the four algorithms in the three environments are depicted in Fig. \ref{fig:result}, where the test case closest to the average outcome is selected. 
The exploration progress of various algorithms at different time points/path lengths is illustrated in Figure \ref{fig:images}. The experimental results demonstrate that our algorithm exhibits superior performance across various stages of exploration completion, particularly excelling during the delicate exploration stage after the majority of the mapping is complete. Compared to four other methods, our algorithm achieves a higher map completion rate at the same time points and requires a shorter exploration path length to attain the same map completion rate. This indicates that our algorithm effectively enhances the efficiency of the exploration process, enabling the robot to make more informed decisions about its actions. However, upon natural termination, our algorithm does show a slight deficiency in map integrity, which may be attributed to our conservative strategy for handling boundary points, primarily adopted to ensure the safety of the exploration process.

In the last column of the Table \ref{table:result}, we recorded whether there were failures during the testing process. We marked exploration processes that did not reach the essential completion level as failures. During the evaluation, our proposed algorithm and boundary exploration methodology consistently performed without failure. Conversely, the RRT algorithm faced challenges in environments Env-2 and Env-4, with the robotic agent becoming trapped in cul-de-sacs. This issue is hypothesized to be due to the RRT algorithm's inadequate management of boundary points. Additionally, the TARE algorithm had a certain probability of exploration failure in all three test environments. Specifically, in Env-2, the exploration was limited to one side of an elongated corridor, and in Env-3, the algorithm only charted the interior of a suite. Most notably, in Env-4, there were consistently rooms left unexplored as the agent retreated to the origin and ceased exploration, with no single trial achieving the essential completion level. We surmise that this behavior may originate from the algorithm's approach to globally guiding updates to local regions. While this global guidance strategy typically bolsters the efficiency of exploration, it occasionally results in incomplete mapping by disregarding essential details, such as the specifics of narrow corridors, thereby failing to explore areas beyond these confined spaces. DA-SLAM \cite{alcalde2022slam}, a raw sensor based DRL exploration algorithm, processes real-time sensor data and current map coverage percentage to generate discrete motion commands. While it demonstrates robust performance in compact environments by efficiently learning navigation policies, its reliance on local perceptual inputs proves detrimental in large-scale settings. Without explored map context, DA-SLAM fails to extrapolate unexplored regions beyond its 8-meter sensing range. This limitation induces cyclical navigation patterns (e.g., prolonged in-place rotation), often culminating in collisions or step-limited terminations. Consequently, DA-SLAM achieves only average $34.24\%$ map coverage in expansive environments, falling short of the essential completion.

\subsection{Ablation Study}
\subsubsection{Effectiveness of Structured Map Representation}
\begin{figure}
	\centering         
	\includegraphics[width=0.9\linewidth]{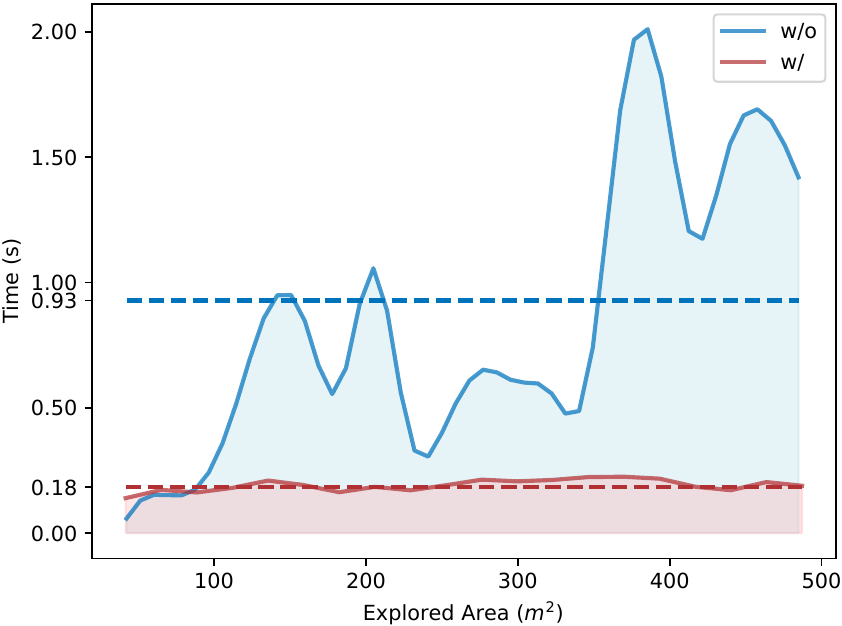} 
	\caption{\small Comparison of processing time between the structured map representation proposed in this paper and direct extraction method for boundary point detection.}
	\label{fig:comparison}
\end{figure}

The structured map representation is instrumental in simplifying data complexity while preserving the essential features of the map. Given the large-scale complex environments under consideration, failing to perform map scaling before subsequent operations would substantially impact processing speed. To illustrate, we conducted a comparative analysis between our structured map representation approach and direct extraction of boundary points, as depicted in Fig. \ref{fig:comparison}. We found that our structured map representation method is fast and stable, while direct extraction is noticeably slower, with increasing time consumption as the explored area expands. Consequently, the structured map representation proves to be an effective strategy for enhancing algorithmic efficiency and processing speed.

\subsubsection{Effectiveness of Action Optimization Unit}

\begin{figure}
	\centering         
	\includegraphics[width=\linewidth]{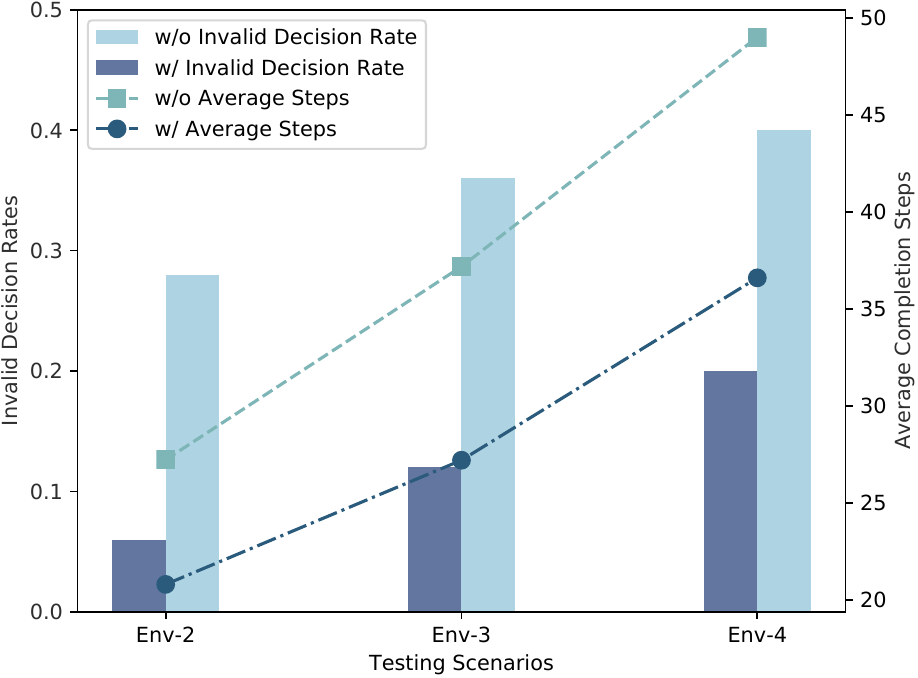} 
	\caption{Effectiveness of AOU.}
	\label{fig:AOU}
\end{figure}

Fig. \ref{fig:AOU} presents ablation results for AOU, demonstrating its exploration enhancement. In Env2-Env4, AOU reduces invalid decisions (unreachable target points) by $22\%$ on average and decreases completion steps by $25\%$. This stems from its boundary optimization mechanism: when the decision network outputs a target in unknown regions, AOU substitutes it with the nearest feasible boundary point. By improving reachability, the method prioritizes regions with higher information density, steering agents away from impractical paths and toward under-explored areas. Consequently, AOU enhances exploration efficiency through targeted, actionable goal selection while minimizing wasted steps on unattainable objectives.

\subsubsection{Impact of ROI size}
\begin{figure}
	\centering         
	\includegraphics[width=\linewidth]{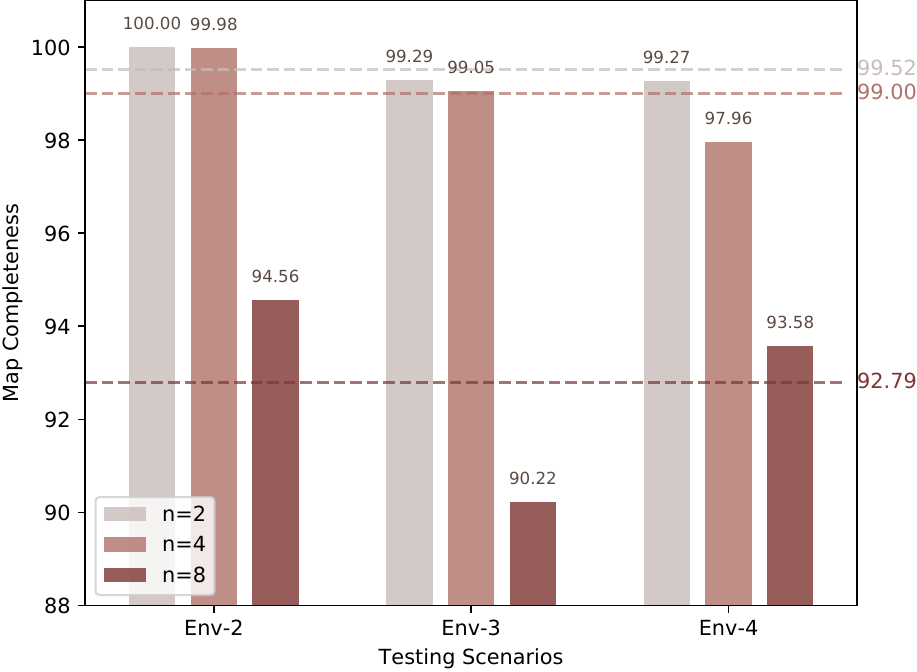} 
	\caption{The impact of different ROI sizes.}
	\label{fig:ROI}
\end{figure}

Our study investigates how ROI size affects exploration outcomes. As shown in Fig. \ref{fig:ROI}, larger ROIs improve exploration efficiency but reduce map completeness. This trade-off arises from diminished available map data as ROI expands, limiting feasible target identification and degrading mapping accuracy. ROI selection must balance scenario area, computational constraints, and completeness requirements. In our tests, setting parameter $n=4$ optimally reconciled efficiency and completeness, ensuring sufficient target points while maintaining systematic exploration capability. 

\subsection{Real-world Experiment}
\begin{figure}
	\centering
	\begin{subfigure}[b]{0.23\textwidth}
		\centering
		\includegraphics[width=\textwidth]{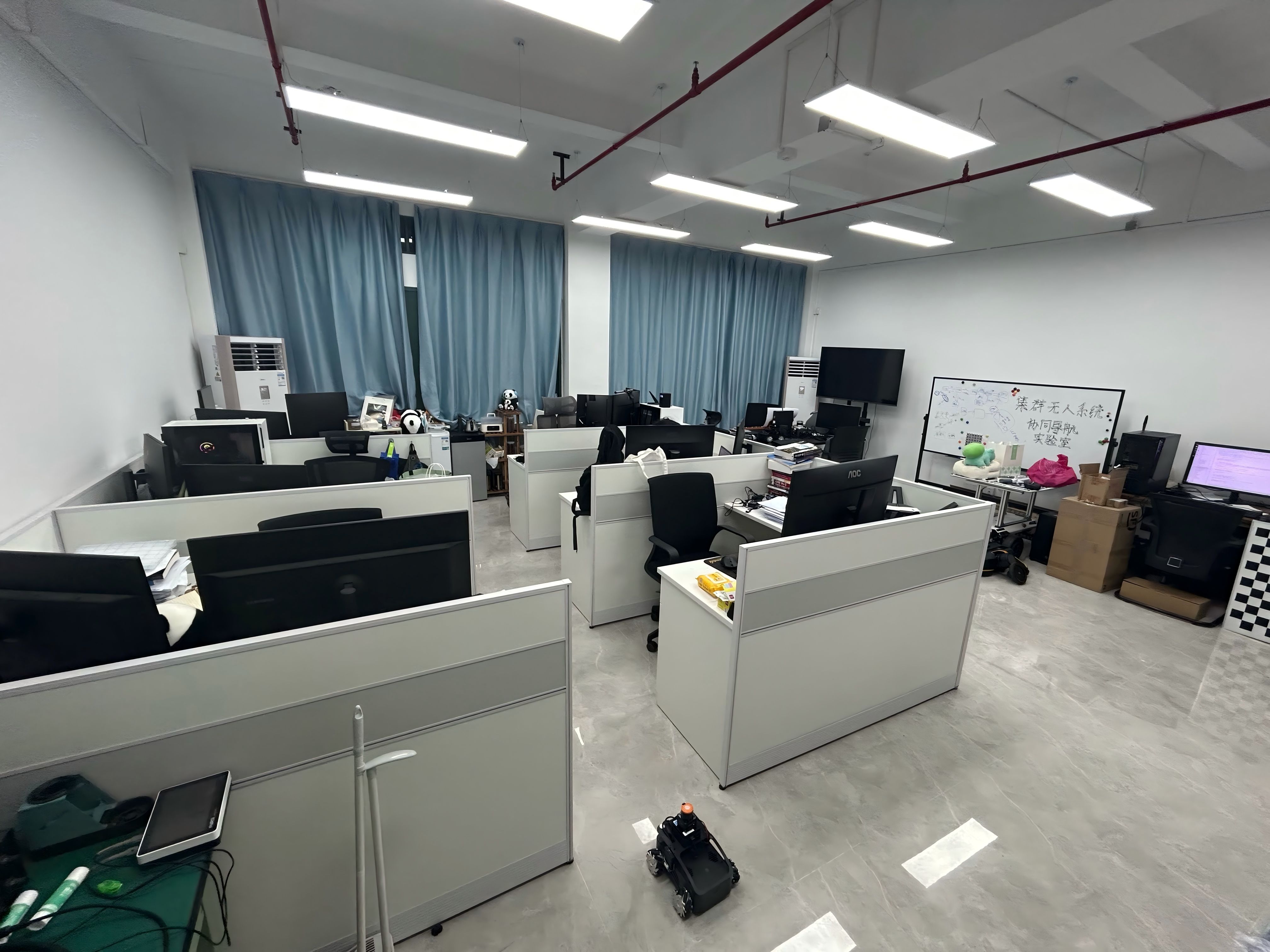}
	\end{subfigure}
	\begin{subfigure}[b]{0.23\textwidth}
		\centering
		\includegraphics[width=\textwidth]{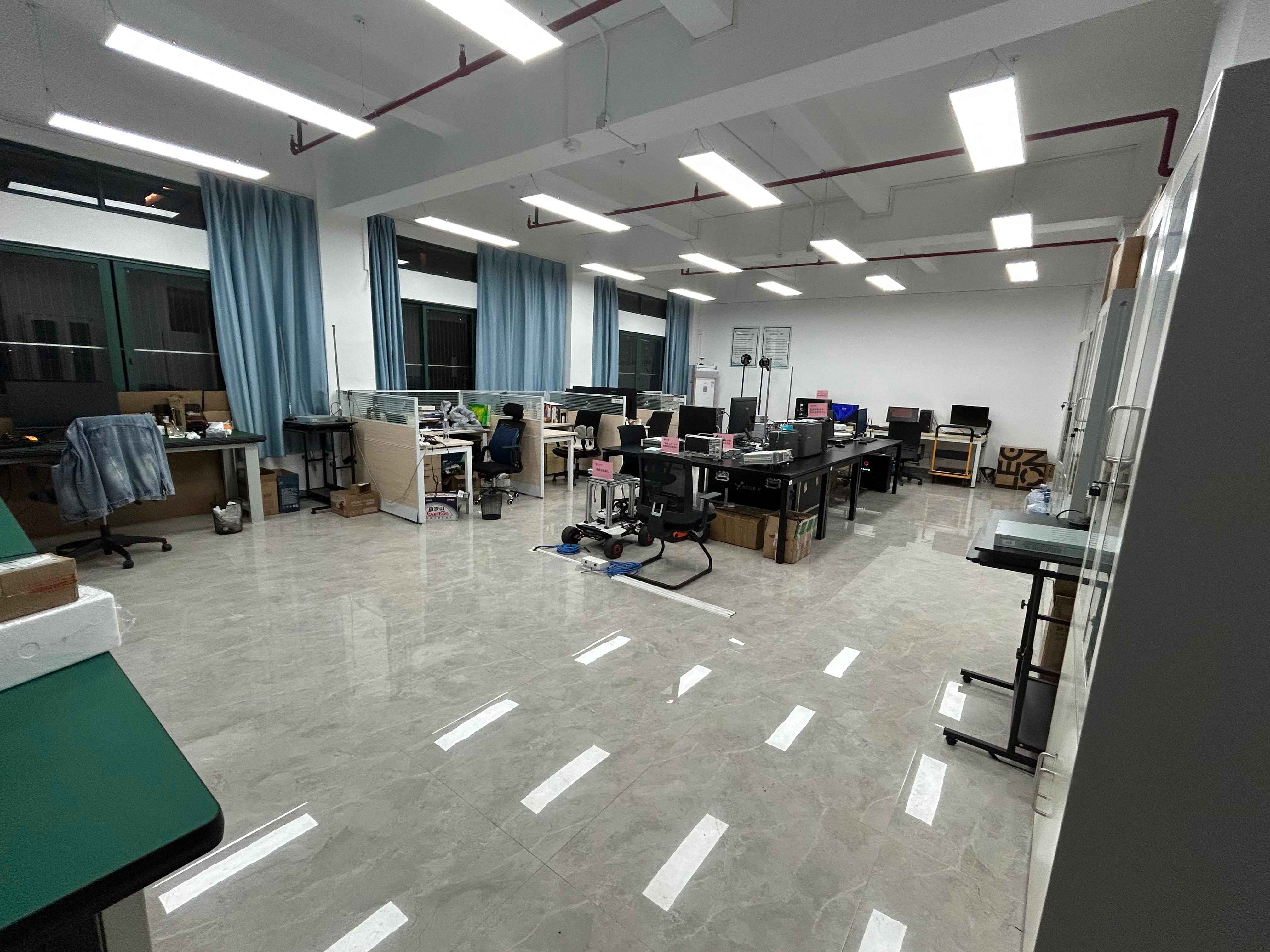}
	\end{subfigure}
	\begin{subfigure}[b]{0.23\textwidth}
		\centering
		\includegraphics[width=\textwidth]{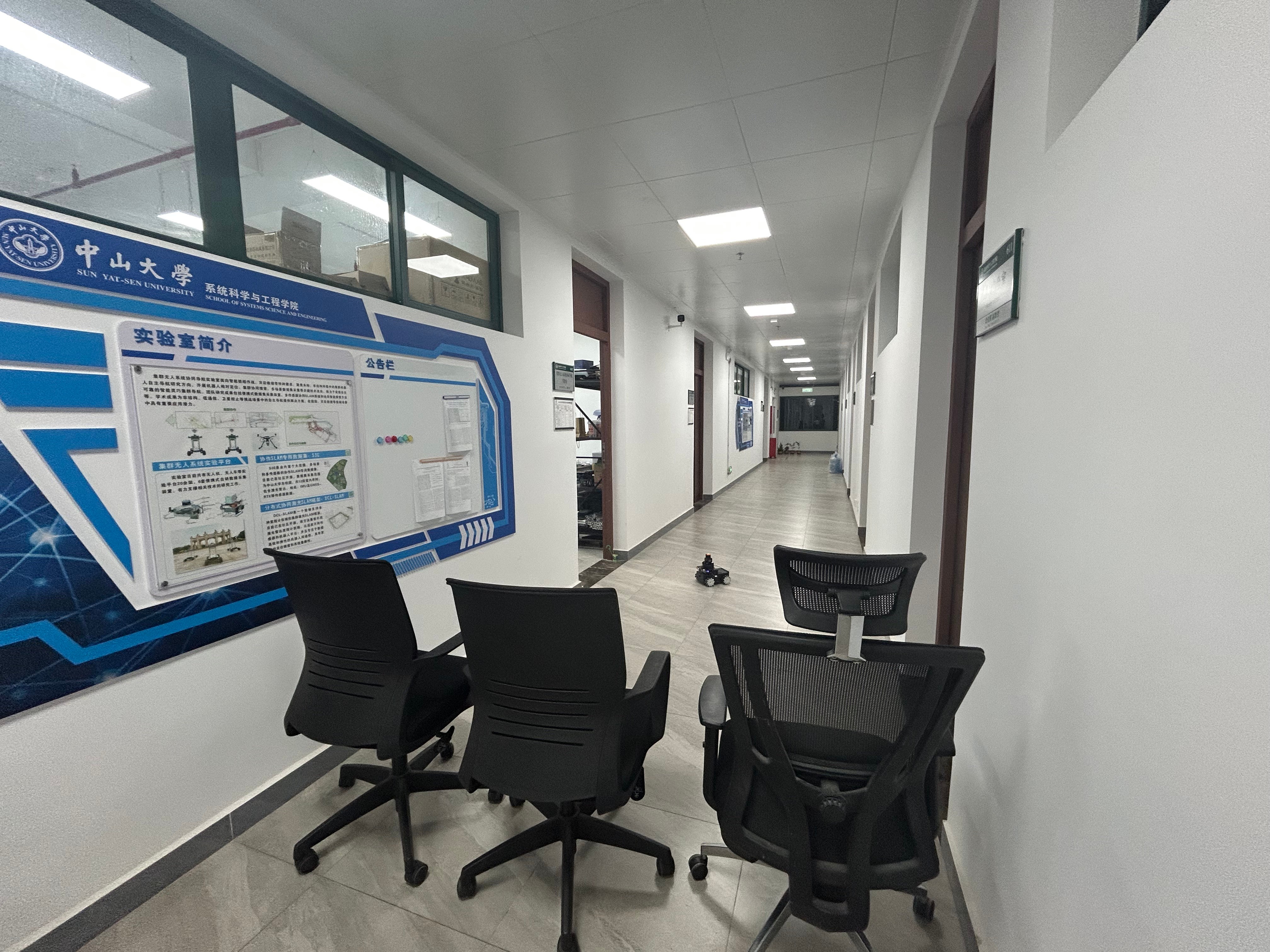}
	\end{subfigure}
	\begin{subfigure}[b]{0.23\textwidth}
		\centering
		\includegraphics[width=\textwidth]{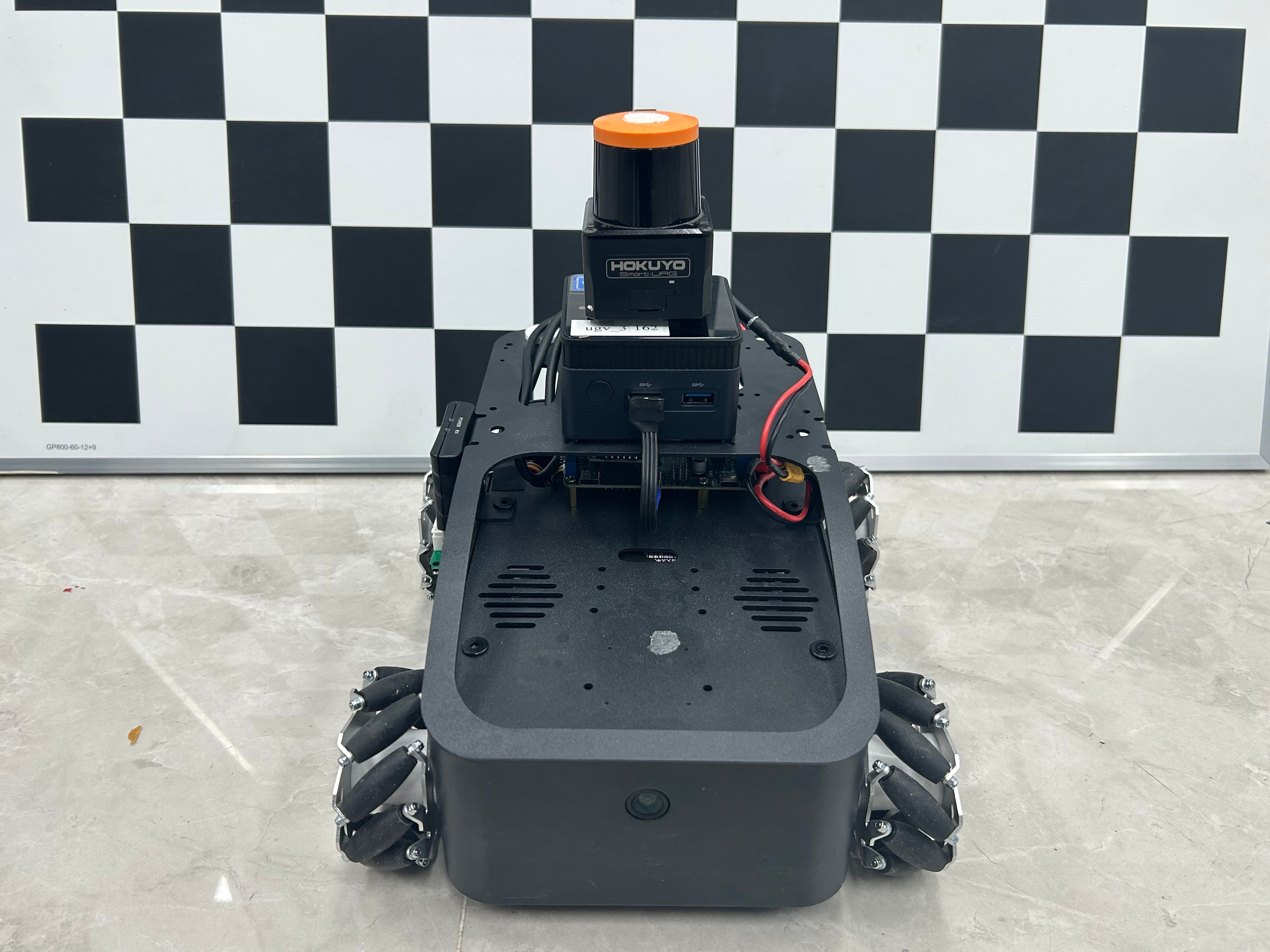}
	\end{subfigure}
	\caption{Real-World Environment.}
	\label{fig:real}
\end{figure}

\begin{figure}
	\centering         
	\includegraphics[width=0.95\linewidth]{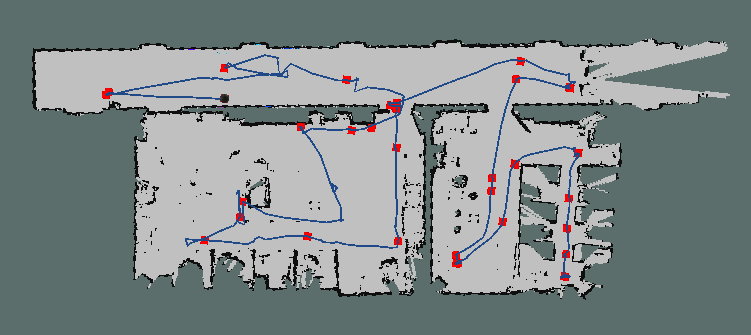} 
	\caption{The real-world constructed map displays red waypoint dots and a blue exploration trajectory curve.}
	\label{fig:realresult}
\end{figure}

We validated our exploration system in a real office environment ($20m\times10m$) using a UGV with Hokuyo UST-10LX lidar and Intel NUC11PH computer (Fig. \ref{fig:real}). The setup, featuring narrow corridors and cluttered obstacles, maintained simulation parameters: $0.45 m/s$ max speed, $8m$ lidar range, and $0.3s$ average decision time (real-time compliant). Over 826 seconds, the UGV navigated 85 meters, achieving near-complete coverage despite some inaccessible zones (Fig. \ref{fig:realresult}). The results demonstrate robust autonomy in complex spaces, with system responsiveness and environmental adaptability balancing exploration efficiency and operational constraints, confirming practical viability beyond simulated scenarios.

\section{Conclusion}
This paper proposes a cutting-edge framework for active SLAM in large-scale complex environments. The framework features a new structured map representation that boosts efficiency by 5 times in map information extraction and boundary point detection. A PPO-based decision-making network is developed, showing a 20\% improvement in exploration efficiency and a 15\% reduction in path length compared to existing methods. The framework is integrated into the Gazebo simulator with realistic robot, sensor, and environment models, and a lightweight SLAM algorithm for real-time autonomous map construction. Furthermore, we deploy a UGV and validate our proposed method in a real office environment. Trained agents can autonomously select optimal actions and apply knowledge to new maps, fulfilling the requirements for effective active SLAM.

\addtolength{\textheight}{-1cm}   




\bibliographystyle{IEEEtran}
\bibliography{ref}

\begin{IEEEbiography}[{\includegraphics[width=1in,height=1.25in,clip,keepaspectratio]{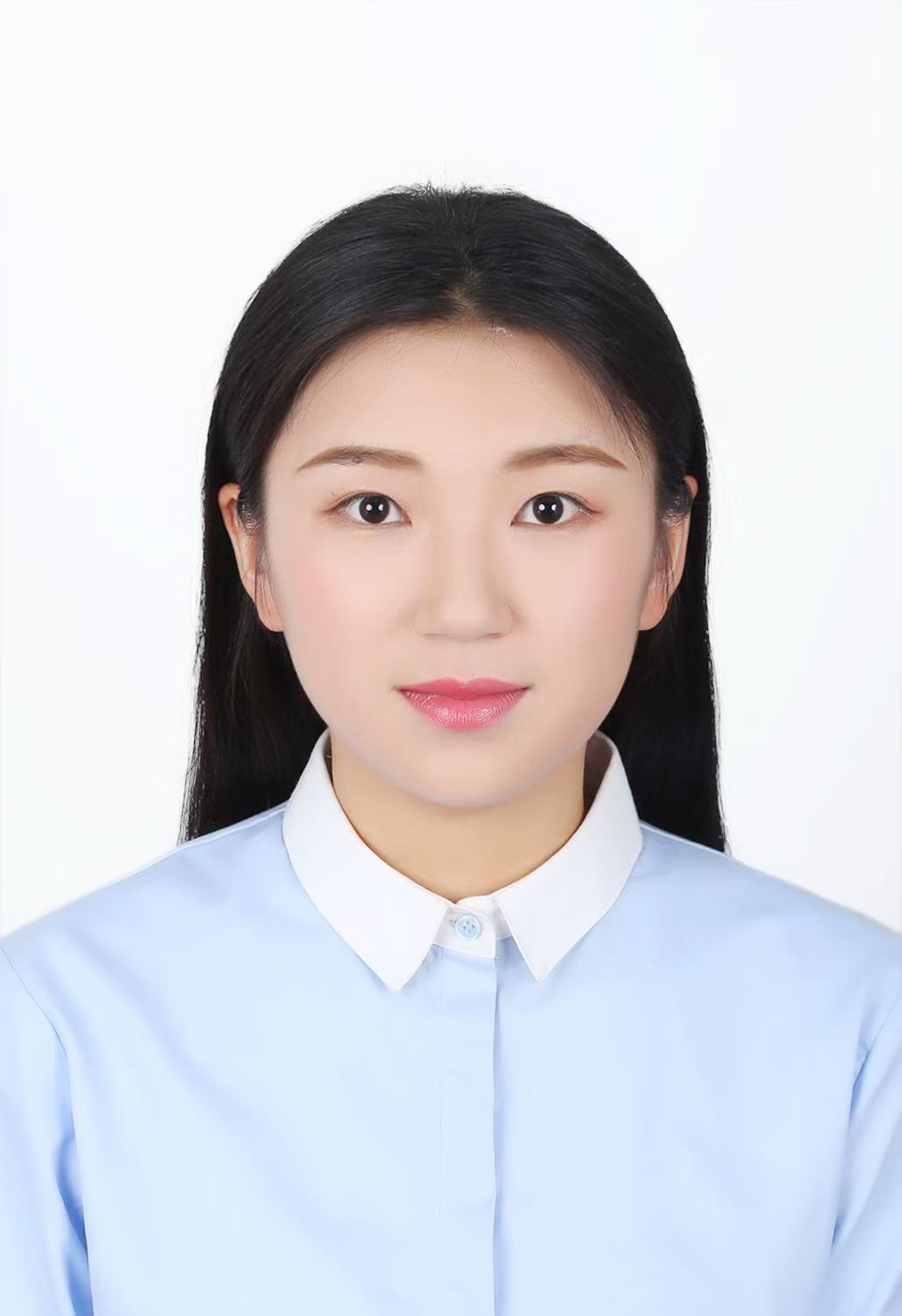}}]{Yizhen Yin}
is a Ph.D. candidate in Computer Science and Technology, Sun Yat-sen University, Guangzhou, China. Previously, She received the M.S. degree in control theory and control engineering from Northeastern University, Shenyang, China. Her research interests include Simultaneous Localization and Mapping (SLAM), Active SLAM, and Path Planning.
\end{IEEEbiography}

\begin{IEEEbiography}[{\includegraphics[width=1in,height=1.25in,clip,keepaspectratio]{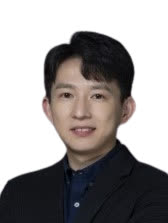}}]{Yuhua Qi}
received the B.S. and Ph.D. degrees in aerospace engineering from the Beijing Institute of Technology, Beijing, China, in 2020. He is currently an Associate Professor at the School of Systems Science and Engineering, Sun Yat-sen University, Guangzhou, China. His research interests include multi-robot simultaneous localization and mapping (SLAM), cooperative control, and autonomous unmanned systems.
\end{IEEEbiography}

\begin{IEEEbiography}[{\includegraphics[width=1in,height=1.25in,clip,keepaspectratio]{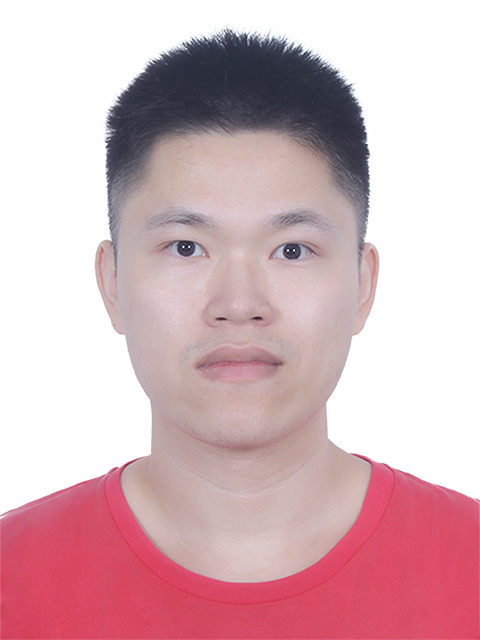}}]{Dapeng Feng}
received the B.E. degree in Computer Science and Technology from Guangdong University of Foreign Studies, Guangzhou, China, in 2018 and the M.S. degree in Pattern Recognition and Intelligent Systems with Sun Yat-sen University, Guangzhou, China, in 2021. He is currently pursuing the Ph.D. degree in Computer Science and Technology with Sun Yat-sen University, Guangzhou, China. He has published several cutting-edge projects on computer vision and robotics, including 3d object detection and SLAM.
\end{IEEEbiography}

\begin{IEEEbiography}[{\includegraphics[width=1in,height=1.25in,clip,keepaspectratio]{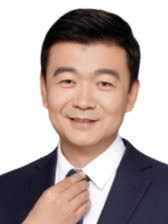}}]{Hongbo Chen}
	received the Ph.D. degree from Harbin Institute of Technology, Harbin, China, in 2007.He is currently a Professor with the School of Systems Science and Engineering, Sun Yat-sen University, Guangzhou, China, where he is the Dean
	of the School of Systems Science and Engineering. His main research interests include modeling, simulation, and analysis of complex systems; intelligent unmanned systems; and design of aerospace vehicles.
\end{IEEEbiography}

\begin{IEEEbiography}[{\includegraphics[width=1in,height=1.25in,clip,keepaspectratio]{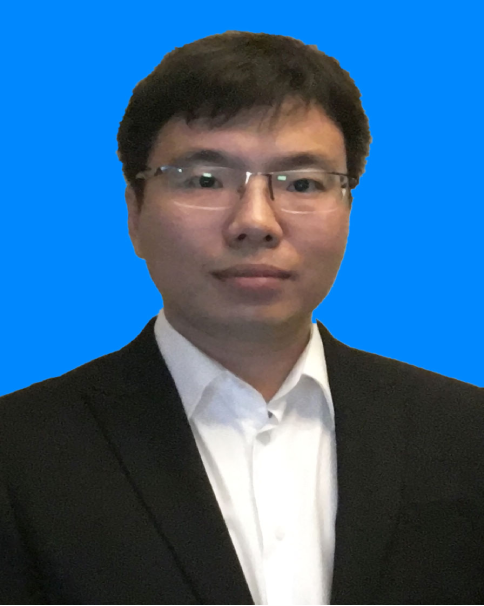}}]{Hongjun Ma}
(Member, IEEE) received the B.S. degree in automation and the Ph.D. degree in control theory and control engineering from Northeastern University, Shenyang, China, in 2004 and 2011, respectively. He is currently a Professor with the School of Automation Science and Engineering, South China University of Technology, Guangzhou, China. His research interests include environmental sensing, decision-making, interactive operation, cluster cooperation, and their applications in intelligent system.
\end{IEEEbiography}

\begin{IEEEbiography}[{\includegraphics[width=1in,height=1.25in,clip,keepaspectratio]{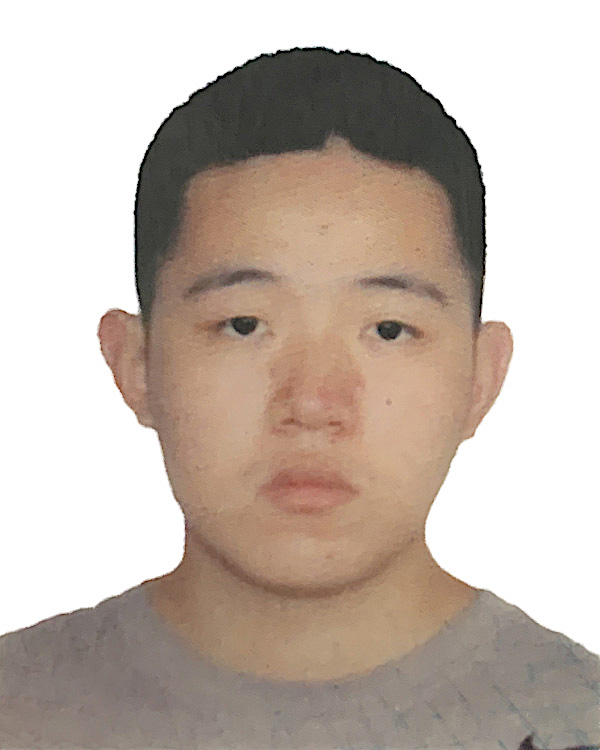}}]{Jin Wu}
	(Member, IEEE) received the B.S. degree from University of Electronic Science and Technology of China, Chengdu, China. He is currently pursuing the Ph.D. degree with The Hong Kong University of Science and Technology, Hong Kong SAR, China. Till now, his related research works have been reported in over 140 journal and conference papers, which include top-tier publications, such as T-RO, T-CAS, T-MECH, T-ASE, T-CYB, and T-AES. His research falls within the algorithmic foundations for robotics with particular focuses on attitude/pose estimation and related mechatronic systems design/control, numerical optimization, deep learning, and high-performance computing.
\end{IEEEbiography}

\begin{IEEEbiography}[{\includegraphics[width=1in,height=1.25in,clip,keepaspectratio]{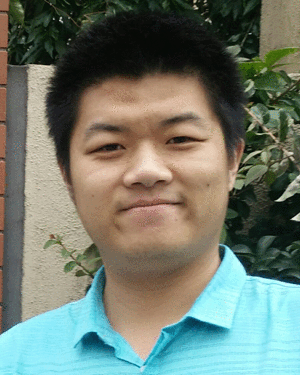}}]{Yi Jiang}
	 (Member, IEEE) received the B.Eng. degree in automation, the M.S. and Ph.D. degrees in control theory and control engineering from Information Science and Engineering College and State Key Laboratory of Synthetical Automation for Process Industries, Northeastern University, Shenyang, China, in 2014, 2016, and 2020, respectively. In 2017, he was a Visiting Scholar with the UTA Research Institute, University of Texas at Arlington, TX, USA, and from 2018 to 2019, he was a Research Assistant with the University of Alberta, Edmonton, Canada. Currently, he is a Postdoc Fellow with the City University of Hong Kong, China. His research interests include networked control systems, industrial process operational control, reinforcement learning, and event-triggered control. Dr. Jiang is an Associate Editor for Advanced Control for Applications: Engineering and Industrial Systems and the recipient of Excellent Doctoral Dissertation Award from Chinese Association of Automation (CAA) in 2021 and Hong Kong Research Grants Council (RGC) Postdoctoral Fellowship Scheme (PDFS) 2023/2024.	
\end{IEEEbiography}

\end{document}